\documentclass[11pt]{article}

\usepackage[final]{acl}

\usepackage{times}
\usepackage{latexsym}

\usepackage{booktabs}   % \toprule, \midrule, \bottomrule, \cmidrule (better horizontal rules)
\usepackage{multirow}   % \multirow (for the Benchmark cell spanning two rows)
\usepackage{makecell}
\usepackage{amsmath}
\usepackage{array}
\renewcommand{\arraystretch}{1.15}
\usepackage{colortbl}
\usepackage{subcaption}
\usepackage{placeins}

\usepackage{hyperref}

\usepackage{array}

\usepackage[T1]{fontenc}
\usepackage[utf8]{inputenc}

\usepackage{microtype}

\usepackage{inconsolata}

\usepackage{graphicx}

\title{System-Mediated Attention Imbalances \\ Make Vision-Language Models Say Yes}

\author{
 \textbf{Tsan Tsai Chan\textsuperscript{1}},
 \textbf{Varsha Suresh\textsuperscript{1}},
 \textbf{Anisha Saha\textsuperscript{1,2}},
 \textbf{Michael Hahn\textsuperscript{1}},
\\
 \textbf{Vera Demberg\textsuperscript{1.2}}
\\
\\
 \textsuperscript{1}Saarland Informatics Campus, Saarland University, Germany,\\
 \textsuperscript{2}Max Planck Institute for Informatics, Germany\\
 {\small\parbox{\textwidth}{
 \centering
   \href{mailto:tsch00001@stud.uni-saarland.de}{tsch00001@stud.uni-saarland.de}, \href{mailto:vsuresh@lst.uni-saarland.de}{vsuresh@lst.uni-saarland.de}, \href{mailto:ansaha@mpi-inf.mpg.de}{ansaha@mpi-inf.mpg.de}, \\ \href{mailto:mhahn@lst.uni-saarland.de}{mhahn@lst.uni-saarland.de}, \href{mailto:vera@lst.uni-saarland.de}{vera@lst.uni-saarland.de}
 }}
}

\begin{document}
\maketitle
\begin{abstract}
Vision-language model (VLM) hallucination is commonly linked to imbalanced allocation of attention across input modalities: system, image and text. However, existing mitigation strategies tend towards an image-centric interpretation of these imbalances, often prioritising increased image attention while giving less consideration to the roles of the other modalities. In this study, we evaluate a more holistic, system-mediated account, which attributes these imbalances to functionally redundant system weights that reduce attention to image and textual inputs. We show that this framework offers a useful empirical perspective on the yes-bias, a common form of hallucination in which VLMs indiscriminately respond `yes'. Causally redistributing attention from the system modality to image and textual inputs substantially suppresses this bias, often outperforming existing approaches. We further present evidence suggesting that system-mediated attention imbalances contribute to the yes-bias by encouraging a default reliance on coarse input representations, which are effective for some tasks but ill-suited to others. Taken together, these findings firmly establish system attention as a key factor in VLM hallucination and highlight its potential as a lever for mitigation. The code for our main interventions can be found at: \url{https://github.com/anisha0325/vlm-hallucination-yes-bias/tree/main}.

\end{abstract}

\section{Introduction}
\label{sec:intro}

For mechanistic analysis, the inputs of vision-language models (VLMs) are generally taken to comprise three modalities: system, image and text \cite{chen2024image, yangunderstanding}.\footnote{We focus on adapter-based VLMs, where all three modalities are passed to the model in a fixed template. Loosely following \citet{chen2024image}, the system modality encompasses everything before the image tokens, including the <BOS> token and system message. The image modality comprises the image patches output by the image encoder and projected into the textual embedding space of the text decoder. The textual modality only consists of the user query for our short-answer benchmarks.} VLM hallucination has been interpreted as resulting from these modalities not being attended to in a balanced way, most often manifesting as insufficient image attention \citep[cf.][]{liu2024paying, shu2025largevisionlanguagemodelalignment, yangunderstanding, zheng2025modality}. This has motivated many existing mitigation strategies to focus primarily on increasing attention to the image modality \cite{liu2024paying, zhu2025ibd, liu2025seeingbelievingprobingdisconnect, sarkar2025mitigatinghallucinationsvisionlanguagemodels, wang2025mllm, zhou2025mitigating, jiang2025devils}. As such, these approaches implicitly assume image attention to be the most important lever against hallucination. We refer to this position as the \textbf{image-centric hypothesis}.

However, in sidelining the other two modalities, image-centric approaches do not provide a holistic account of attention imbalances. The system modality, in particular, contains substantial amounts of functionally redundant attention in the form of attention sinks \cite{xiao2024efficient, gu2025when}: tokens which receive disproportionate attention despite being semantically uninformative. Prior work reports that this modality often accounts for over 70\% of the total attention in VLM decoders~\cite{chen2024image, tang2025intervening, yangunderstanding, liang2026not}. This concentration suggests that system attention may play a central role in hallucination by contributing to attention imbalances, a possibility prior work has not rigorously examined. 

To address this gap, we investigate an alternative, more holistic view of hallucination, which we term the \textbf{system-mediated hypothesis}. Instead of focussing only on image attention, this hypothesis posits that hallucination arises from the joint influence of two factors: one, \textit{redundant} system attention and two, the resulting \textit{insufficient} attention to both the image and textual modalities. We demonstrate that a system-mediated account is empirically superior to an image-centric one for a widespread form of VLM hallucination, the yes-bias. This form of single-token hallucination involves VLMs responding `yes' in binary yes/no settings irrespective of the prompt~\cite{zhang2016yin, ross2024what, li2023evaluating, li2024naturalbench, zhang2025illusionbench}. It occurs in a diverse range of binary tasks, most notably object detection \citep{li2023evaluating, zheng2025modality}, reasoning \citep{zhang2025illusionbench} and factual question-answering \citep{li2024naturalbench}. %Affecting binary settings, the yes-bias can be objectively evaluated against clear-cut ground-truth answers.

Using causal interventions that redistribute attention weights between modalities in different parts of a VLM's text decoder, we find that redundant late-layer system attention plays a decisive role in promoting the yes-bias by causing localised attention imbalances. Consistent with the system-mediated hypothesis, correcting these imbalances by redistributing attention to the image and textual modalities substantially suppresses the bias and outperforms alternative image- and text-centric mitigation strategies. These gains do not exclusively depend on increasing image attention, undermining the universality of the image-centric hypothesis. We then present evidence suggesting that such late-layer attention imbalances contribute to the yes-bias by encouraging a default reliance on coarse input representations, which is detrimental for certain tasks.

Our contributions are twofold:

(1) We identify redundant late-layer system attention as an important contributor to the yes-bias, highlighting the system modality's relevance to hallucination.

(2) Using principled redistributions of attention weights, we formalise a system-mediated account of VLM hallucination as a more holistic alternative to prevailing image-centric interpretations of imbalanced attention. This motivates attention-based mitigation methods that do not focus on one modality in isolation.

\section{Related Work}
\label{sec:rel_w}

\subsection{Attention-based hallucination mitigation in VLMs} 

Despite their wide acceptance \citep{liu2024paying, liu2025seeingbelievingprobingdisconnect, wang2025mllm, sarkar2025mitigatinghallucinationsvisionlanguagemodels, zheng2025modality}, mitigation strategies that increase image attention typically posit that insufficient image attention is the foremost contributor to hallucination without objectively measuring the scale of the presupposed attention deficit. 
%This prevents any success in suppressing hallucination from being straightforwardly equated to a confirmation of the image-centric hypothesis. 
Any success in suppressing hallucination is then simplistically equated to a confirmation of the image-centric hypothesis.
Reliance on arbitrary hyperparameters is common as a result, for example in scaling image weights by coefficients with questionable generalisability \citep[cf.][]{liu2024paying}.

More fundamentally, this body of work has not shown that non-image modalities' attention patterns contribute \textit{less} to VLM hallucination. This is a non-trivial omission, as increasing either the absolute magnitude or relative prominence of image attention weights implies reducing the influence of the other input modalities. Our targeted causal interventions directly address this gap by rigorously contrasting the trade-offs entailed by changes to the attention weights of each modality.

\subsection{Role of system attention}
Although prior work widely presumes that removing or redistributing weights from attention sinks suppresses VLM hallucination \citep{huang2024opera, kang2025see}, the large attention sinks in the system modality discussed above have yet to be strategically exploited for hallucination mitigation. Instead, the modality as a whole has generally been dismissed as irrelevant to hallucination \citep{bi2025unveiling, yangunderstanding} or lumped together with the textual one \citep{sarkar2025mitigatinghallucinationsvisionlanguagemodels}. While \citet{liang2026not} examine the functions of attention sinks within the system modality by zero-ablating their attention weights, the study largely neglects the potential causal role of such tokens in VLM hallucination. Our system-mediated interpretation enables us to directly investigate just this by testing a clear hypothesis: that large system weights induce hallucination through depriving both the image and textual modalities of attention.

\subsection{The yes-bias in VLMs}

Studies on the causes of VLM hallucination have chiefly evaluated models and interventions on a combination of yes/no --- i.e.~binary, mostly single-token generation --- and autoregressive tasks, e.g.~image captioning \citep{bi2025unveiling, gong-etal-2024-damro, huang2024opera, shu2025largevisionlanguagemodelalignment, liu2025reducing, zhou2025mitigating, DBLP:journals/corr/abs-2503-06169}. Although intended to test generalisability, this has in practice conflated hallucinatory behaviour in different settings and overlooks evidence that yes/no and autoregressive hallucination may have distinct underlying causes (\citealt{zheng2025modality}, also cf. \citealt{parcalabescu2025do}).

As a result, the mechanistic causes of the yes-bias have not been adequately investigated beyond tentative references to imbalanced training data \citep{hu2023ciem, liu2024survey, he2025evaluating, hu2025prescribing}. This neglect is difficult to justify as the yes-bias crops up in a wide variety of binary tasks and is specifically targeted by many commonly used benchmarks, including POPE \citep{li2023evaluating} and HallusionBench \citep{Guan_2024_CVPR}. It is likewise a missed opportunity to isolate clean mechanistic signals of hallucination that can be intervened on. This comes about because yes/no tasks permit objective evaluation in contrast to autoregressive settings, where hallucination-related signals are more ambiguous \citep{huang2024opera} and evaluation often relies on less objective methods such as GPT-as-judge \citep[e.g.][]{yue2024mmmu, ghosh2025visual, liu2025reducing}.

Collectively, the above makes the case for causal interventions on a VLM's text decoder to evaluate the feasibility of a system-mediated account of the yes-bias against an image-centric one. The next section details how we implement these.

\section{Causal Intervention Framework}
\label{sec:framework}
\subsection{Disambiguating image-centric and system-mediated hypotheses}
\label{subsec:disambiguating}
Our causal interventions are designed to differentiate the image-centric and system-mediated hypotheses through their effectiveness in suppressing the yes-bias. To operationalise these two hypotheses, we define a modality to have insufficient attention if increasing attention weights enhances performance. Conversely, a modality has functionally redundant attention if ablating the attention weights of all its tokens either does not degrade performance or leads to improvement.

To reiterate, the image-centric hypothesis assumes that hallucination arises mainly from insufficient image attention. This view accordingly predicts that the yes-bias is best mitigated by increasing attention to the image. In contrast, our system-mediated hypothesis attributes hallucination to functionally redundant system attention weights inducing deficits in both the image and textual modalities. As such, it predicts that redistributing attention from the system modality to boost attention to the image and/or textual modalities will be most effective.

\subsection{Proportional redistribution of attention weights}
\label{subsec:proportional}

As a minimal test of both hypotheses, we design an intervention to probe the role of each modality in producing the yes-bias. Specifically, we apply what we term \textit{proportional redistribution} to the system and textual modalities in turn. Each of these serves as \textit{source modalities}. From these, we reallocate \textit{all} attention weights to the remaining two modalities, the \textit{recipient modalities},\footnote{Unlike \citet{yangunderstanding}, we perform redistributions instead of just rescaling modality-specific weights without renormalisation for numerical stability.} in proportion to the latter's original attention weights.

This approach operationalises the predictions outlined in Section~\ref{subsec:disambiguating}. If the system-mediated hypothesis is correct, proportionally redistributing attention from the system modality towards both the image and textual modalities should produce the largest performance gains. Conversely, under the image-centric hypothesis, proportional redistributions from either the system or textual modalities should yield comparable improvements as both would increase image attention.

Formally, we let $\mathcal{M} = {\text{system}, \text{image}, \text{text}}$ represent the set of input modalities. For each modality $m \in \mathcal{M}$, let $A_i$ denote the post-softmax attention weight assigned to token $i \in m$, aggregated across heads and decoder query positions within the decoder layers we intervene on. The total attention mass assigned to modality $m$ is then defined as:

\begin{equation}
\alpha_m \;=\; \sum_{i \in m} A_i,
\label{eq:modality_mass}
\end{equation}

Further, let $s \in \{\text{system}, \text{text}\}$ denote the source modality. The set of recipient modalities is then defined as $R = \mathcal{M} \setminus \{s\}$, comprising the two remaining modalities. On this basis, proportional redistribution is implemented as a two-step process: first, zero-ablation of the source modality's attention weights (Equation~\ref{eq:source_zero}) and second, reallocation of the ablated weights to the recipient modalities in proportion to their original attention weights (Equation~\ref{eq:proportional_redistribution}).

\vspace{-0.5em}
\begin{equation}
\alpha^{\prime}_{s} \;=\; 0,
\label{eq:source_zero}
\end{equation}
\vspace{-0.5em}
\vspace{-0.5em}
\begin{equation}
\alpha^{\prime}_{r} \;=\; \alpha_{r}
\;+\;
\alpha_{s}
\cdot
\frac{\alpha_{r}}{\sum_{r' \in R} \alpha_{r'}},
\quad
r \in R,
\label{eq:proportional_redistribution}
\end{equation}
\vspace{-0.5em}

\section{Main Experiments}
\label{sec:experiment}
\subsection{Evaluation set-up}
\noindent\textbf{\textit{Architectures}}. Our interventions focus on the text decoder of LLaVA-1.5 7B \citep{liu2023visual} as it is widely used in interpretability studies and to test hallucination-mitigating techniques. Appendix \ref{app:other} reports supplementary results from interventions on additional models from the same family, LLaVA-NeXT-Vicuna 7B \citep{liu2024llavanext} and LLaVA-1.5 13B.  The applicability of our conclusions to Qwen2-VL-Instruct \citep{team2024qwen2}, which does not belong to the same family, is also critically evaluated in Appendix \ref{app:qwen2}.

\noindent\textbf{\textit{Scope of interventions}}.
Prior work situates model components responsible for hallucination in the later layers of the text decoder \citep{DBLP:journals/corr/abs-2404-18930, chen2024multiobject, wang2025mllm}. Accordingly, we partition the 32-layer text decoder of LLaVA-1.5 7B into four quarters, each consisting of eight attention layers. We refer to these as Q1-4 and focus on Q4 (layers 25 to 32), taking it to encompass the decoder's late layers.\footnote{Quartering the model is a compromise as per-head and layer interventions are computationally infeasible. Given our still developing understanding of how VLM functions are spread across decoder layers, segmenting the model this way is arguably more principled than the customary practice of dividing it into early, middle and late segments, each comprising different numbers of layers \citep{bi2025unveiling, neo2025towards, wang2025mllm}.} Appendix \ref{app:late_layer} shows that our focus on Q4 --- rather than other model components --- is empirically well founded.

\begin{table*}[t]
  \centering
  \footnotesize
  \renewcommand{\arraystretch}{1.05}
  \begin{tabular}{p{0.28\textwidth} p{0.47\textwidth} p{0.15\textwidth}}
    \toprule
    \textbf{Benchmark} & \textbf{Purpose} & \textbf{No. of prompts} \\
    \midrule
    BEAF \citep{yebin2024beaf} &
    Adversarial object detection (objects digitally edited out) &
    26{,}000 \\

    HallusionBench \citep{Guan_2024_CVPR} &
    Various (e.g.\ optical illusions, OCR, reasoning) &
    951 \\

    MME \citep{fu2025mme} &
    Various, with emphasis on knowledge extraction &
    2{,}374 \\

    NaturalBench \citep{li2024naturalbench} &
    Verification of the same captions against minimally different images &
    5{,}800 (yes/no subset) \\
    
    SugarCrepe \citep{hsieh2023sugarcrepe} &
    Verification of minimally different captions against the same image &
    1{,}820 (subset) \\
    
    Winoground \citep{thrush2022winoground} &
    Verification of captions only differing in word order against images &
    5{,}664 \\
    \bottomrule
  \end{tabular}
  \caption{\label{tab:benchmarks}
    Overview of the six yes/no paired-prompt vision-language benchmarks used in this study.
  }
\end{table*}

\noindent\textbf{\textit{Benchmarks}}. To investigate the degree of yes-bias with and without intervention, we perform evaluations on six benchmarks specifically designed for this purpose (Table \ref{tab:benchmarks}). Each benchmark consists of paired prompts that differ minimally from each other either in their image or in the user query such that the ground-truth answer is `yes' for one prompt and `no' for the other. Where this is not already enforced, we explicitly request yes/no answers in the user prompts.

\noindent\textbf{\textit{Evaluation metrics}}. We use three complementary metrics to quantify the yes-bias:

(1) \textbf{Simple accuracy}: Proportion of prompts answered correctly

(2) \textbf{Paired accuracy}: Proportion of prompt pairs with both answers correct

(3) \textbf{Yes-rate}: Proportion of `yes' out of the total number of responses generated, ignoring correctness \citep[cf.][]{Guan_2024_CVPR}

The difference between the yes-rate and the actual proportion of ground-truth-yes prompts is our most direct measure of the yes-bias following \citet{Guan_2024_CVPR}. Simple and paired accuracy quantify its impact on performance: default `yes' responses harm simple accuracy and are especially penalised by paired accuracy given the paired-prompt design of our benchmarks. Unless otherwise stated, we report only simple accuracy and exclude benchmark-specific metrics (e.g.~Accuracy+ for MME).

\subsection{The system-mediated hypothesis is valid}
\label{subsec:val}

\begin{table*}
\centering
\footnotesize
\renewcommand{\arraystretch}{1.15}
\begin{tabular}{
p{0.11\textwidth}
p{0.09\textwidth}
p{0.08\textwidth}
p{0.08\textwidth}
p{0.08\textwidth}
p{0.08\textwidth}
p{0.08\textwidth}
p{0.08\textwidth}
p{0.08\textwidth}
}
\toprule
\textbf{Benchmark (\% prompts with ground-truth yes)} & \textbf{Metric} & \textbf{PAI} & \textbf{Image$\times$ 2.0} & \textbf{AD-HH} & \textbf{Q4 text redistr (prop)} & \textbf{Q4 system abl} & \textbf{Q4 system redistr (prop)} & \textbf{No-intervention baseline} \\
\midrule

\multirow{3}{=}{BEAF \\ (34.11\% yes)}
& \makecell[l]{Simple acc$^{\Uparrow}$}
& \makecell[l]{83.01\\(–0.87)}
& \makecell[l]{85.73\\(+2.38)}
& \makecell[l]{83.67\\(-0.08)}
& \makecell[l]{85.55\\(+2.16)}
& \makecell[l]{84.44\\(–3.76)}
& \makecell[l]{\textbf{88.40}\\(+5.56)}
& 83.74 \\

& \makecell[l]{Paired acc$^{\Uparrow}$}
& \makecell[l]{77.72\\(–1.14)}
& \makecell[l]{81.01\\(+3.04)}
& \makecell[l]{78.51\\(–0.14)}
& \makecell[l]{80.75\\(+2.71)}
& \makecell[l]{79.45\\(+1.05)}
& \makecell[l]{\textbf{84.74}\\(+7.78)}
& 78.62 \\

& \makecell[l]{Yes-rate$^{\Delta\downarrow}$}
& \makecell[l]{45.73\\(+34.07)}
& \makecell[l]{41.55\\(+21.81)}
& \makecell[l]{44.74\\(+31.16)}
& \makecell[l]{41.68\\(+22.19)}
& \makecell[l]{43.58\\(+27.77)}
& \makecell[l]{\textbf{32.62}\\(–4.37)}
& \makecell[l]{44.69\\(+31.02)} \\

\midrule
\multirow{3}{=}{HallusionBench (42.17\% yes)}
& \makecell[l]{Simple acc$^{\Uparrow}$}
& \makecell[l]{45.64\\(0.00)}
& \makecell[l]{45.95\\(+0.68)}
& \makecell[l]{45.22\\(–0.92)}
& \makecell[l]{46.58\\(+2.06)}
& \makecell[l]{45.74\\(+0.22)}
& \makecell[l]{\textbf{53.31}\\(+16.81)}
& 45.64 \\

& \makecell[l]{Paired acc$^{\Uparrow}$}
& \makecell[l]{9.53\\(–2.46)}
& \makecell[l]{10.00\\(+2.35)}
& \makecell[l]{9.30\\(–4.81)}
& \makecell[l]{9.77\\(0.00)}
& \makecell[l]{10.00\\(+2.35)}
& \makecell[l]{\textbf{19.07}\\(+95.19)}
& 9.77 \\

& \makecell[l]{Yes-rate$^{\Delta\downarrow}$}
& \makecell[l]{90.22\\(+113.94)}
& \makecell[l]{88.22\\(+109.20)}
& \makecell[l]{90.43\\(+114.44)}
& \makecell[l]{84.23\\(+99.74)}
& \makecell[l]{89.27\\(+111.70)}
& \makecell[l]{\textbf{59.62}\\(+41.38)}
& \makecell[l]{89.80\\(+112.95)} \\

\midrule
\multirow{3}{=}{MME \\ (50.00\% yes)}
& \makecell[l]{Simple acc$^{\Uparrow}$}
& \makecell[l]{78.73\\(–0.32)}
& \makecell[l]{78.35\\(–0.80)}
& \makecell[l]{78.73\\(–0.32)}
& \makecell[l]{78.26\\(–0.92)}
& \makecell[l]{78.90\\(–0.11)}
& \makecell[l]{72.83\\(–8.44)}
& \textbf{78.98} \\

& \makecell[l]{Paired acc$^{\Uparrow}$}
& \makecell[l]{\textbf{30.77}\\(0.00)}
& \makecell[l]{\textbf{30.77}\\(0.00)}
& \makecell[l]{29.23\\(–5.00)}
& \makecell[l]{29.23\\(–5.00)}
& \makecell[l]{30.00\\(–2.50)}
& \makecell[l]{21.54\\(–30.00)}
& \textbf{30.77} \\

& \makecell[l]{Yes-rate$^{\Delta\downarrow}$}
& \makecell[l]{42.62\\(-14.76)}
& \makecell[l]{40.14\\(–19.72)}
& \makecell[l]{\textbf{42.88}\\(-14.24)}
& \makecell[l]{41.07\\(–17.86)}
& \makecell[l]{41.20\\(–17.61)}
& \makecell[l]{28.98\\(–42.04)}
& \makecell[l]{42.29\\(-15.42)} \\

\midrule
\multirow{3}{=}{NaturalBench (50.00\% yes)}
& \makecell[l]{Simple acc$^{\Uparrow}$}
& \makecell[l]{59.97\\(–0.42)}
& \makecell[l]{61.28\\(+1.76)}
& \makecell[l]{59.79\\(–0.71)}
& \makecell[l]{62.17\\(+3.24)}
& \makecell[l]{60.97\\(+1.24)}
& \makecell[l]{\textbf{66.02}\\(+9.63)}
& 60.22 \\

& \makecell[l]{Paired acc$^{\Uparrow}$}
& \makecell[l]{21.10\\(–2.09)}
& \makecell[l]{23.72\\(+10.07)}
& \makecell[l]{20.66\\(–4.13)}
& \makecell[l]{25.52\\(+18.42)}
& \makecell[l]{23.00\\(+6.73)}
& \makecell[l]{\textbf{34.59}\\(+60.51)}
& 21.55 \\

& \makecell[l]{Yes-rate$^{\Delta\downarrow}$}
& \makecell[l]{86.03\\(+72.06)}
& \makecell[l]{83.83\\(+67.66)}
& \makecell[l]{86.34\\(+72.68)}
& \makecell[l]{82.07\\(+64.14)}
& \makecell[l]{84.55\\(+69.10)}
& \makecell[l]{\textbf{66.67}\\(+33.34)}
& \makecell[l]{85.67\\(+71.34)} \\

\midrule
\multirow{3}{=}{SugarCrepe (50.00\% yes)}
& \makecell[l]{Simple acc$^{\Uparrow}$}
& \makecell[l]{57.47\\(–0.10)}
& \makecell[l]{58.18\\(+1.13)}
& \makecell[l]{57.14\\(–0.68)}
& \makecell[l]{58.79\\(+2.19)}
& \makecell[l]{57.97\\(+0.76)}
& \makecell[l]{\textbf{62.75}\\(+9.07)}
& 57.53 \\

& \makecell[l]{Paired acc$^{\Uparrow}$}
& \makecell[l]{15.05\\(–0.73)}
& \makecell[l]{16.37\\(+7.98)}
& \makecell[l]{14.40\\(–5.01)}
& \makecell[l]{17.58\\(+15.96)}
& \makecell[l]{16.04\\(+5.83)}
& \makecell[l]{\textbf{25.93}\\(+71.04)}
& 15.16 \\

& \makecell[l]{Yes-rate$^{\Delta\downarrow}$}
& \makecell[l]{91.98\\(+83.96)}
& \makecell[l]{91.37\\(+82.74)}
& \makecell[l]{92.53\\(+85.06)}
& \makecell[l]{90.66\\(+81.32)}
& \makecell[l]{91.59\\(+83.19)}
& \makecell[l]{\textbf{85.71}\\(+71.42)}
& \makecell[l]{92.03\\(+84.06)} \\

\midrule
\multirow{3}{=}{Winoground (50.00\% yes)}
& \makecell[l]{Simple acc$^{\Uparrow}$}
& \makecell[l]{53.50\\(+0.22)}
& \makecell[l]{53.69\\(+0.58)}
& \makecell[l]{53.06\\(–0.60)}
& \makecell[l]{55.00\\(+3.03)}
& \makecell[l]{53.56\\(+0.34)}
& \makecell[l]{\textbf{57.88}\\(+8.43)}
& 53.38 \\

& \makecell[l]{Paired acc$^{\Uparrow}$}
& \makecell[l]{7.63\\(+3.39)}
& \makecell[l]{8.85\\(+19.92)}
& \makecell[l]{6.63\\(–10.16)}
& \makecell[l]{12.00\\(+62.60)}
& \makecell[l]{8.13\\(+10.09)}
& \makecell[l]{\textbf{19.50}\\(+164.23)}
& 7.38 \\

& \makecell[l]{Yes-rate$^{\Delta\downarrow}$}
& \makecell[l]{95.25\\(+90.5)}
& \makecell[l]{94.56\\(+89.12)}
& \makecell[l]{95.94\\(+91.88)}
& \makecell[l]{92.00\\(+84.00)}
& \makecell[l]{94.81\\(+89.63)}
& \makecell[l]{\textbf{75.38}\\(+50.76)}
& \makecell[l]{95.38\\(+90.76)} \\

\bottomrule
\end{tabular}
\caption{\label{tab:interventions} A comparison of different interventions on simple accuracy, paired accuracy and yes-rate across six paired-prompt benchmarks. Best results for each metric are in bold. Our simple redistribution of system attention weights (Q4 system redistr (prop)) consistently outperforms the image and text-based alternatives by a sizeable margin on all three metrics in five of the six benchmarks. Moreover, simply removing Q4 system attention without proportional redistribution (Q4 system ablation) does not bring about the sharp improvements in performance that Q4 system redistribution does. For simple and paired accuracy, percentage (not percentage-point) differences relative to the no-intervention baseline are shown in brackets. The differences for yes-rate are with reference to the actual proportion of prompts with ground truth ‘yes, shown bracketed in the Benchmark column --- the best yes-rate is the one that deviates least from that ground-truth value.}
\end{table*}

We first apply proportional redistribution of Q4 system and text attention as described in Section \ref{subsec:proportional}. As a shorthand, we refer to these as \textit{Q4 system} and \textit{Q4 text redistribution}, respectively. To verify the need for positing two factors under our system-mediated account of hallucination (see Section \ref{sec:intro}), we also perform Q4 system ablation. This removes system attention without redistributing or renormalising weights across modalities, enabling us to distinguish the effects of removing system attention from those of redistributing it to the other modalities.

We compare these interventions against a no-intervention baseline, interventions that boost image attention \citep[PAI,][and $\text{Image}\times2.0$]{liu2024paying} and a method that suppresses text attention like Q4 text redistribution \citep[AD-HH, ][]{yangunderstanding}. Additional implementation details for these alternative methods are provided in Appendix~\ref{app:baseline_prop}.

\noindent\textbf{\textit{Results}}. Compared to image-boosting and text-suppressing alternatives, Q4 system redistribution mitigates the yes-bias much more effectively on our three metrics and in five of our six benchmarks (Table \ref{tab:interventions}). The severity of this bias is particularly evident in HallusionBench, SugarCrepe and Winoground, where the no-intervention baselines exhibit yes-rates exceeding 90\% despite ground-truth yes proportions being closer to 50\%. It is for these benchmarks that Q4 system redistribution produces the largest improvements in paired accuracy, which more than doubles for Winoground.

Within Q4, these findings confirm the system-mediated hypothesis' prediction, i.e.~removing system weights to boost both image and text attention effectively reduces the yes-bias. MME exhibits a different pattern, which we analyse in Section \ref{sec:error_analysis}. As a sanity check, Appendix \ref{app:other_responses} further presents evidence that such interventions affect yes/no responses much more strongly than other types of generation, justifying our view of the yes-bias as a distinct form of VLM hallucination.

Likewise, Q4 system ablation clearly indicates the necessity of such a two-factor account of hallucination. Without also increasing both image and text attention, removing system attention fails to bring about substantially better performance, yielding only negligible improvements (again Table \ref{tab:interventions}). In sum, these results affirm the empirical soundness of our system-mediated hypothesis as operationalised in our set-up.

\subsection{The image-centric hypothesis is insufficient}
\label{sec:pairwise}
\begin{table*}[htbp]
\centering
\footnotesize
\begin{tabular}{l|ccc|ccc|ccc}
\toprule
\multirow{3}{*}{\textbf{Source}} & \multicolumn{9}{c}{\textbf{Recipient}} \\
\cline{2-10}
& \multicolumn{3}{c}{Simple acc$^{\Uparrow}$} & \multicolumn{3}{c}{Paired acc$^{\Uparrow}$} & \multicolumn{3}{c}{Yes-rate$^{\Delta\downarrow}$ (ground truth 42.17\%)} \\
\cline{2-10}
& \textit{System} & \textit{Image} & \textit{Text} & \textit{System} & \textit{Image} & \textit{Text} & \textit{System} & \textit{Image} & \textit{Text} \\
\midrule
\textit{System} & \cellcolor{lightgray}45.64 & \makecell{\underline{51.52} \\ (+12.88)} & \makecell{\textbf{58.04} \\ (+27.17)} & \cellcolor{lightgray}9.77 & \makecell{\underline{15.12} \\ (+54.76)} & \makecell{\textbf{23.26} \\ (+138.08)} & \cellcolor{lightgray}\makecell{89.80 \\ (+112.95)} & \makecell{\underline{65.82} \\ (+56.08)} & \makecell{\textbf{39.12} \\ (--7.23)} \\[1ex]
\textit{Image} & \makecell{45.43 \\ (--0.46)} & \cellcolor{lightgray}45.64 & \makecell{45.53 \\ (--0.24)} & \makecell{9.30 \\ (--4.81)} & \cellcolor{lightgray}9.77 & \makecell{9.53 \\ (--2.46)} & \makecell{88.75 \\ (+110.46)} & \cellcolor{lightgray}\makecell{89.80 \\ (+112.95)} & \makecell{89.7 \\ (+112.71)} \\[1ex]
\textit{Text} & \makecell{46.69 \\ (+2.30)} & \makecell{47.21 \\ (+3.44)} & \cellcolor{lightgray}45.64 & \makecell{10.00 \\ (+2.35)} & \makecell{10.23 \\ (+4.71)} & \cellcolor{lightgray}9.77 & \makecell{84.12 \\ (+99.48)} & \makecell{83.39 \\ (+97.75)} & \cellcolor{lightgray}\makecell{89.80 \\ (+112.95)} \\[1ex]
\bottomrule
\end{tabular}
\caption{Changes in performance in HallusionBench under pairwise attention-weight transfers between all combinations of source and recipient modalities in Q4 of LLaVA-1.5 7B. System-to-text transfers consistently yield the biggest improvements in performance, followed by system-to-image transfers. Performance is measured in simple and paired accuracy alongside yes-rate, with percentage differences from the no-intervention baseline (simple/ paired accuracy) or the ground truth (yes-rate) indicated in brackets. The best score for each metric is bolded, the second best underlined and baseline configurations are marked by shading.}
\label{tab:pairwise}
\end{table*}

\begin{table*}[h]
\centering
\footnotesize
\begin{tabular}{l|ccc|ccc|ccc|ccc}
\toprule
\multirow{3}{*}{\textbf{Source}} & \multicolumn{12}{c}{\textbf{Recipient}} \\
\cline{2-13}
& \multicolumn{3}{c}{\textbf{BEAF}} 
& \multicolumn{3}{c}{\textbf{NaturalBench (yes/no)}} 
& \multicolumn{3}{c}{\textbf{SugarCrepe}} 
& \multicolumn{3}{c}{\textbf{Winoground}} \\
\cline{2-13}
& \textit{System} & \textit{Image} & \textit{Text}
& \textit{System} & \textit{Image} & \textit{Text}
& \textit{System} & \textit{Image} & \textit{Text}
& \textit{System} & \textit{Image} & \textit{Text} \\
\midrule
\textit{System} 
& \cellcolor{lightgray}82.38 & \underline{88.06} & \textbf{88.08}
& \cellcolor{lightgray}60.22 & \underline{65.59} & \textbf{67.50}
& \cellcolor{lightgray}57.53 & \underline{62.47} & \textbf{64.95}
& \cellcolor{lightgray}53.38 & \underline{57.19} & \textbf{58.44} \\[1ex]

\textit{Image} 
& 84.17 & \cellcolor{lightgray}82.38 & 83.71
& 60.60 & \cellcolor{lightgray}60.22 & 60.10
& 57.74 & \cellcolor{lightgray}57.53 & 57.53
& 53.63 & \cellcolor{lightgray}53.38 & 53.31 \\[1ex]

\textit{Text} 
& 85.56 & 85.76 & \cellcolor{lightgray}82.38
& 62.10 & 62.47 & \cellcolor{lightgray}60.22
& 58.79 & 58.96 & \cellcolor{lightgray}57.53
& 55.06 & 54.88 & \cellcolor{lightgray}53.38 \\[1ex]
\bottomrule
\end{tabular}
\caption{Changes in simple accuracy in BEAF, NaturalBench (yes/no), SugarCrepe, and Winoground under pairwise attention-weight transfers between all combinations of source and recipient modalities in Q4 of LLaVA-1.5 7B. System-to-text transfers consistently yield the biggest improvements in performance, followed by system-to-image transfers. The best score is bolded, the second best underlined and baseline configurations are marked by shading.}
\label{tab:pairwise-2}
\end{table*}

\begin{table}[h]
\centering
\footnotesize
\begin{tabular}{l|ccc}
\toprule
\multirow{3}{*}{\textbf{Source}} & \multicolumn{3}{c}{\textbf{Recipient}} \\
\cline{2-4}
& \multicolumn{3}{c}{\textbf{MME}} \\
\cline{2-4}
& \textit{System} & \textit{Image} & \textit{Text} \\
\midrule
\textit{System} 
& \cellcolor{lightgray}78.98 & 66.39 & 61.25 \\[1ex]

\textit{Image} 
& 78.85 & \cellcolor{lightgray}78.98 & 78.48 \\[1ex]

\textit{Text} 
& 78.22 & 78.18 & \cellcolor{lightgray}78.98 \\[1ex]
\bottomrule
\end{tabular}
\caption{Changes in simple accuracy in MME under pairwise attention-weight transfers between all combinations of source and recipient modalities in Q4 of LLaVA-1.5 7B. Exactly contrary to the other benchmarks, the system-to-text transfer consistently yields the \textit{worst} performance, followed by the system-to-image transfer.}
\label{tab:pairwise-mme}
\end{table}

To more cleanly evaluate the predictions of both hypotheses as set out in Section \ref{subsec:disambiguating}, we next examine the relative impact of selectively boosting image versus text attention. For this purpose, we perform \textit{pairwise} redistributions that transfer removed weights in Q4 to only one recipient modality rather than splitting them between two. This yields transfers in both directions between the system and image, system and text, and image and text modalities. Based on the account above, the image-centric hypothesis predicts that image-boosting transfers (system-to-image and text-to-image) will suppress the yes-bias more effectively than text-boosting ones (here, system-to-text). The system-mediated hypothesis, by contrast, predicts system-suppressing transfers (system-to-image and system-to-text) will be superior to image-boosting ones (text-to-image).

\noindent\textit{\textbf{Results}}. Overall, the greatest improvements in all three metrics are obtained with system-to-text transfers, followed by the system-to-image and text-to-image settings. Table \ref{tab:pairwise} illustrates that for HallusionBench. The pattern generalises to the remaining benchmarks (Table \ref{tab:pairwise-2}), again with MME as the only exception (Table \ref{tab:pairwise-mme}). With system-to-text transfers outperforming both image-boosting interventions, the prediction of our system-mediated hypothesis is validated. At the same time, the gains brought about by boosting text attention show a strong formulation of the image-centric hypothesis --- that the image modality is the only one with an attention deficit --- to be untenable in our set-up.

Collectively, these findings imply a substantial attention imbalance in Q4, characterised by redundant attention allocated to system tokens and insufficient attention directed towards both image and text tokens, rather than image tokens alone. 

\section{Error Analysis}
\label{sec:error_analysis}

\subsection{Distinguishing coarse- from fine-grained tasks}

\begin{table*}[!h]
\centering
\footnotesize
\renewcommand{\arraystretch}{1.15}
\begin{tabular}{
p{0.12\textwidth}
p{0.09\textwidth}
p{0.09\textwidth}
p{0.09\textwidth}
p{0.09\textwidth}
p{0.09\textwidth}
p{0.10\textwidth}
p{0.09\textwidth}
}
\toprule
\textbf{POPE split} &
\textbf{Metric} &
\textbf{PAI} &
\textbf{Image$\times$2.0} &
\textbf{AD-HH} &
\textbf{Q4 text (prop)} &
\textbf{Q4 system (prop)} &
\textbf{No-intervention baseline} \\
\midrule

\multirow{2}{=}{Overall \\ (50.00\% yes)}
& \makecell[l]{Simple acc$^{\Uparrow}$}
& \makecell[l]{86.56\\(–0.35)}
& \makecell[l]{\textbf{87.67}\\(+0.92)}
& \makecell[l]{86.77\\(–0.12)}
& \makecell[l]{87.56\\(+0.79)}
& \makecell[l]{87.20\\(+0.38)}
& 86.87 \\

& \makecell[l]{Yes-rate$^{\Delta\downarrow}$}
& \makecell[l]{54.00\\(+8.00)}
& \makecell[l]{\textbf{50.53}\\(+1.06)}
& \makecell[l]{53.19\\(+6.38)}
& \makecell[l]{50.71\\(+1.42)}
& \makecell[l]{43.49\\(–13.02)}
& \makecell[l]{53.22\\(+6.44)} \\
\midrule

\multirow{2}{=}{Adversarial \\ (50.00\% yes)}
& \makecell[l]{Simple acc$^{\Uparrow}$}
& \makecell[l]{81.73\\(–0.61)}
& \makecell[l]{83.67\\(+1.75)}
& \makecell[l]{82.03\\(–0.24)}
& \makecell[l]{83.43\\(+1.45)}
& \makecell[l]{\textbf{85.10}\\(+3.49)}
& 82.23 \\

& \makecell[l]{Yes-rate$^{\Delta\downarrow}$}
& \makecell[l]{58.87\\(+17.74)}
& \makecell[l]{54.53\\(+9.06)}
& \makecell[l]{57.97\\(+15.94)}
& \makecell[l]{54.83\\(+9.66)}
& \makecell[l]{\textbf{45.50}\\(–9.00)}
& \makecell[l]{57.90\\(+15.80)} \\
\midrule

\multirow{2}{=}{Popular \\ (50.00\% yes)}
& \makecell[l]{Simple acc$^{\Uparrow}$}
& \makecell[l]{87.50\\(–0.53)}
& \makecell[l]{\textbf{88.60}\\(+0.72)}
& \makecell[l]{87.83\\(–0.16)}
& \makecell[l]{88.47\\(+0.57)}
& \makecell[l]{87.80\\(–0.19)}
& 87.97 \\

& \makecell[l]{Yes-rate$^{\Delta\downarrow}$}
& \makecell[l]{53.03\\(+6.06)}
& \makecell[l]{49.60\\(–0.80)}
& \makecell[l]{52.10\\(+4.20)}
& \makecell[l]{\textbf{49.80}\\(–0.40)}
& \makecell[l]{42.93\\(–14.14)}
& \makecell[l]{52.10\\(+4.20)} \\
\midrule

\multirow{2}{=}{Random \\ (50.00\% yes)}
& \makecell[l]{Simple acc$^{\Uparrow}$}
& \makecell[l]{90.43\\(+0.03)}
& \makecell[l]{90.73\\(+0.37)}
& \makecell[l]{90.43\\(+0.03)}
& \makecell[l]{\textbf{90.77}\\(+0.41)}
& \makecell[l]{88.70\\(–1.88)}
& 90.40 \\

& \makecell[l]{Yes-rate$^{\Delta\downarrow}$}
& \makecell[l]{\textbf{50.10}\\(+0.20)}
& \makecell[l]{47.47\\(–5.06)}
& \makecell[l]{49.50\\(–1.00)}
& \makecell[l]{47.50\\(–5.00)}
& \makecell[l]{42.03\\(–15.94)}
& \makecell[l]{49.67\\(-0.66)} \\

\bottomrule
\end{tabular}
\caption{\label{tab:pope}Shifts in simple accuracy and yes-rate for each split of POPE (MSCOCO) under different interventions. Q4 system redistribution only benefits performance in the adversarial split, while harming it in the two others. Paired accuracy does not apply to this benchmark because while balanced, prompts are not explicitly grouped into positive/negative pairs.}
\end{table*}

Having shown that Q4 system-mediated attention imbalances are major contributors to the yes-bias, we now examine why redistributing this attention yields gains in all our paired-prompt benchmarks except MME. We start with two critical clues: one, Q4 system redistribution suppresses the yes-rate across all benchmarks --- it is only in MME that this fails to translate into performance gains; two, Q4 system ablation does not suppress the yes-rate despite removing system attention (Table \ref{tab:interventions}).

A salient commonality of our non-MME benchmarks is that they explicitly require compositional analysis of multimodal inputs to distinguish between minimally different prompt pairs (see Table \ref{tab:benchmarks}). In other words, these tasks prioritise fine-grained, local detail, and systematically penalise reliance on high-level representations of inputs.

By contrast, MME instantiates a class of tasks where such representations are often appropriate. Many MME prompts involve verifying if images depict specific types of artwork, landmarks, films, or scenes. This arguably requires matching high-level representations of known referents across modalities more than local visual detail. 

To show the relevance of this distinction, we introduce a seventh benchmark, POPE \citep{li2023evaluating}. POPE is a balanced yes/no task that requires models to verify if an object is in an image. It comprises three splits, differing in how absent objects are selected. Q4 system redistribution improves performance only in the adversarial split, while degrading it in the other two despite identical ground-truth yes-rates throughout (Table \ref{tab:pope}).

Crucially, POPE's adversarial split resembles the non-MME benchmarks in that correct responses hinge on identifying subtle differences in objects present, hence necessitating fine-grained scrutiny of visual inputs. Here, absent objects are selected to be related to the scene in the image, making reliance on high-level visual representations actively misleading. The other two splits, on the other hand, do not penalise such reliance as harshly, as absent objects are chosen either randomly or by their dataset frequency. Example prompts from representative benchmarks are included in Appendix \ref{app:examples}.

\subsection{Late-layer system attention and cross-modal aggregation}
This pattern motivates a unified interpretation of the seven benchmarks: system-mediated attention imbalances in the baseline model encourage the reliance on coarse, aggregated representations of its inputs as a learnt default. This reliance reflects attention allocation patterns learnt during training that are helpful for tasks where fine-grained analysis is not crucial, but detrimental where it is.

Mirroring the two factors under our system-mediated hypothesis, we submit that this reliance arises through a dual mechanism. First, attention sinks such as the \texttt{<BOS>} token in the system modality encourage the coarse, cross-modal aggregation of information \cite{barbero2025why}. Second, image and text attention deficits hinder the extraction of fine-grained information from the inputs \cite{bi2025unveiling, wang2025mllm}. When coarse, aggregated representations dominate, responses rely chiefly on high-level information; where this does not suffice for an adequate response, the model falls back on a second default behaviour, the yes-bias \cite{hu2023ciem, liu2024survey, hu2025prescribing}.

Such an interpretation provides a principled explanation for the performance differences between Q4 system ablation and redistribution. As it only addresses aggregation by removing system attention, Q4 system ablation falls short of increasing the use of fine-grained multimodal information, thereby failing to disable the default yes-bias. Q4 system redistribution, on the other hand, both reduces cross-modal aggregation \textit{and} facilitates access to finer-grained image and text information. Provided with information helpful in compositional tasks, the model relies less on default-yes responses, thus suppressing the yes-bias.

\subsection{Behavioural evidence}
LLaVA-1.5 7B's yes-rates support linking coarse representations to default model behaviour. Tasks focussing on such representations exhibit markedly smaller disparities between no-intervention yes-rates and the ground truth than compositional tasks. For MME, this disparity is only 15.42\%, compared to gaps ranging from 31.02\% to 112.95\% for the remaining benchmarks (Table \ref{tab:interventions}). A similar pattern holds among the three POPE splits, where the adversarial split exhibits a 15.80\% disparity, in stark contrast to 4.20\% and 0.66\% respectively for the other two splits (Table \ref{tab:pope}). Consequently, it is precisely the tasks least aligned with default behaviour that benefit from intervention.

\subsection{Representational evidence}

\begin{table*}[t]
  \centering
  \footnotesize
  \renewcommand{\arraystretch}{1.05}
  \begin{tabular}{p{0.13\textwidth} p{0.1\textwidth} p{0.11\textwidth} p{0.11\textwidth} p{0.07\textwidth} p{0.1\textwidth} p{0.1\textwidth} p{0.08\textwidth}}
    \toprule
    \textbf{Dataset} & \textbf{No. (layers $\times$ prompts)} & \textbf{Mean cos sim (no intervention)} & \textbf{Mean cos sim (Q4 redistribution)} & \textbf{Paired t-stat} & \textbf{Holm p (t-test)} & \textbf{Wilcoxon stat} & \textbf{Holm p (Wilcoxon)} \\
    \midrule
    BEAF & 182{,}448 & 0.139 & 0.100 & 762.154 & $<0.001$ & 271{,}170{,}836.5 & $<0.001$ \\

    HallusionBench & 6{,}657 & 0.127 & 0.087 & 129.159 & $<0.001$ & 593{,}527.5 & $<0.001$ \\

    MME & 16{,}618 & 0.136 & 0.090 & 187.214 & $<0.001$ & 4{,}711{,}829.5 & $<0.001$ \\

    POPE & 63{,}000 & 0.139 & 0.095 & 454.681 & $<0.001$ & 30{,}506{,}750 & $<0.001$ \\

    NaturalBench & 46{,}400 & 0.135 & 0.092 & 391.512 & $<0.001$ & 27{,}924{,}061 & $<0.001$ \\

    SugarCrepe & 790{,}622 & 0.136 & 0.092 & 1634.860 & $<0.001$ & 4{,}407{,}350{,}131 & $<0.001$ \\

    Winoground & 11{,}200 & 0.137 & 0.090 & 198.562 & $<0.001$ & 740{,}974 & $<0.001$ \\
    \bottomrule
  \end{tabular}
  \caption{\label{tab:cos_sim}
    Redistributing late-layer (Q4) system attention brings about a clearly lower image-text cosine similarity in the layers affected by intervention. This is shown by the similarity values under intervention (Mean cos sim (Q4 redistribution)) being significantly lower than those without intervention (Mean cos sim (no intervention)) for all seven datasets covered here. All p-values obtained are consistently under 0.001. For each dataset and condition (with or without intervention), cosine similarity values are averaged across Q4 layers and prompts within that dataset.
  }
\end{table*}

There is likewise mechanistic evidence more directly in favour of our account. If indeed late-layer system attention encourages cross-modal aggregation of representations --- i.e. representations that are coarser-grained and more global, we predict that the presence of this attention would make image and text representations more similar. By implication, \textit{removing} this attention should make them \textit{less} similar.

Confirming our predictions, redistributing system attention significantly lowers image-text cosine similarity in late decoder layers across our benchmarks. Using a paired t-test and Wilcoxon sign-rank test, we consistently obtain Holm-corrected p-values of less than 0.001 as shown in Table \ref{tab:cos_sim}. Layers without intervention show no change.

Both these lines of evidence underscore the close association between system-mediated attention imbalances and the yes-bias. An image-centric account, by contrast, fails to predict the specific failure modes of system-based interventions.

\section{Conclusion}
\label{sec:conclusion}
Above, we investigated the yes-bias, a widespread form of VLM hallucination, and showed that attributing it solely to insufficient image attention is empirically inadequate. Our causal interventions on LLaVA-1.5 7B revealed that redundant system attention in late decoder layers drives imbalances that harm both image and text processing, possibly by promoting reliance on coarse input representations. Redistributing this system attention substantially reduces the yes-bias, outperforming existing image- and text-oriented mitigation strategies for compositional tasks.

Building on this, future work should explore the implications of our findings for autoregressive hallucination, complementing recent work relating VLM answers to their explanations \citep{parcalabescu2025do}. It would be especially instructive to apply holistic causal interventions, similar to what we used, to other manifestations of VLM hallucination in order to make out whether hallucination-inducing attention imbalances share any commonalities. Further, given the robust improvements system-to-text redistributions yielded in our set-up, the prospect of redundant system attention driving the yes-bias in text-only LLMs should also be investigated.

\section*{Limitations}

Our work focusses on LLaVA-based architectures, chosen for their widespread use in interpretability studies. We also restrict our scope to single-token generation in yes/no contexts. The generality of our system-mediated account should therefore be tested on a broader variety of VLM families and in a wider range of settings. The latter ought to cover other types of default single-token responses in multiple-choice and counting tasks for instance \citep[cf.][]{li2024naturalbench, sim2025can}, as well as autoregressive generation.

Next, our interventions mostly employ zero-ablation with complete redistribution to avoid arbitrary assumptions about how much attention to transfer between modalities. This has prevented us from methodically exploring the effects of partial redistributions, to investigate for example how monotonic their impacts on performance are. We likewise concentrate on the quarter level owing to computational constraints, precluding a per-layer study analogous to \citet{shi2025vision}.

Finally, our claims regarding representation-level mechanisms should be further validated through directly probing a model's internal states, e.g.~to quantify whether tasks we claim to be more closely aligned with its default state induce less uncertainty than those less well aligned. This would supplement the behavioural and representational evidence we adduced above.

\section*{Acknowledgements}
Varsha Suresh is funded by Deutsche Forschungsgemeinschaft, Funder ID: \url{http://dx.doi.org/10.13039/501100001659}, SFB 1102: 'Information Density and Linguistic Encoding'.

\bibliography{latex/custom}

@inproceedings{
luo2026from,
title={From Narrow to Panoramic Vision: Attention-Guided Cold-Start Reshapes Multimodal Reasoning},
author={Ruilin Luo and Chufan Shi and Yizhen Zhang and Cheng Yang and Songtao Jiang and Tongkun Guan and Ruizhe Chen and Ruihang Chu and Peng Wang and Mingkun Yang and Lei Wang and Yujiu Yang and Junyang Lin and Zhibo Yang},
booktitle={The Fourteenth International Conference on Learning Representations},
year={2026},
url={https://openreview.net/forum?id=4tsfY0lI1w}
}

@InProceedings{Bitton-Guetta_2023_ICCV,
    author    = {Bitton-Guetta, Nitzan and Bitton, Yonatan and Hessel, Jack and Schmidt, Ludwig and Elovici, Yuval and Stanovsky, Gabriel and Schwartz, Roy},
    title     = {Breaking Common Sense: WHOOPS! A Vision-and-Language Benchmark of Synthetic and Compositional Images},
    booktitle = {Proceedings of the IEEE/CVF International Conference on Computer Vision (ICCV)},
    month     = {October},
    year      = {2023},
    pages     = {2616-2627}
}

@article{DBLP:journals/corr/abs-2503-06169,
  publtype={informal},
  author={Shawn Li and Jiashu Qu and Yuxiao Zhou and Yuehan Qin and Tiankai Yang and Yue Zhao},
  title={Treble Counterfactual VLMs: A Causal Approach to Hallucination},
  year={2025},
  month={March},
  cdate={1740787200000},
  journal={CoRR},
  volume={abs/2503.06169},
  url={https://doi.org/10.48550/arXiv.2503.06169}
}

@inproceedings{he2025evaluating,
  title={Evaluating and Mitigating Object Hallucination in Large Vision-Language Models: Can They Still See Removed Objects?},
  author={He, Yixiao and Sun, Haifeng and Ren, Pengfei and Wang, Jingyu and Wang, Huazheng and Qi, Qi and Zhuang, Zirui and Wang, Jing},
  booktitle={Proceedings of the 2025 Conference of the Nations of the Americas Chapter of the Association for Computational Linguistics: Human Language Technologies (Volume 1: Long Papers)},
  pages={6841--6858},
  year={2025}
}

@inproceedings{
tang2025intervening,
title={Intervening Anchor Token: Decoding Strategy in Alleviating Hallucinations for {MLLM}s},
author={Feilong Tang and Zile Huang and Chengzhi Liu and Qiang Sun and Harry Yang and Ser-Nam Lim},
booktitle={The Thirteenth International Conference on Learning Representations},
year={2025},
url={https://openreview.net/forum?id=zGb4WgCW5i}
}

@misc{liu2024llavanext,
    title={LLaVA-NeXT: Improved reasoning, OCR, and world knowledge},
    url={https://llava-vl.github.io/blog/2024-01-30-llava-next/},
    author={Liu, Haotian and Li, Chunyuan and Li, Yuheng and Li, Bo and Zhang, Yuanhan and Shen, Sheng and Lee, Yong Jae},
    month={January},
    year={2024}
}

@inproceedings{zhang2016yin,
  title={Yin and yang: Balancing and answering binary visual questions},
  author={Zhang, Peng and Goyal, Yash and Summers-Stay, Douglas and Batra, Dhruv and Parikh, Devi},
  booktitle={Proceedings of the IEEE conference on computer vision and pattern recognition},
  pages={5014--5022},
  year={2016}
}

@inproceedings{
ross2024what,
title={What makes a good metric? Evaluating automatic metrics for text-to-image consistency},
author={Candace Ross and Melissa Hall and Adriana Romero-Soriano and Adina Williams},
booktitle={First Conference on Language Modeling},
year={2024},
url={https://openreview.net/forum?id=LFfktMPAci}
}

@inproceedings{thrush2022winoground,
  title={Winoground: Probing vision and language models for visio-linguistic compositionality},
  author={Thrush, Tristan and Jiang, Ryan and Bartolo, Max and Singh, Amanpreet and Williams, Adina and Kiela, Douwe and Ross, Candace},
  booktitle={Proceedings of the IEEE/CVF Conference on Computer Vision and Pattern Recognition},
  pages={5238--5248},
  year={2022}
}

@inproceedings{jiang2025devils,
  title={Devils in middle layers of large vision-language models: Interpreting, detecting and mitigating object hallucinations via attention lens},
  author={Jiang, Zhangqi and Chen, Junkai and Zhu, Beier and Luo, Tingjin and Shen, Yankun and Yang, Xu},
  booktitle={Proceedings of the Computer Vision and Pattern Recognition Conference},
  pages={25004--25014},
  year={2025}
}

@inproceedings{
hu2023ciem,
title={{CIEM}: Contrastive Instruction Evaluation Method for Better Instruction Tuning},
author={Hongyu Hu and Jiyuan Zhang and Minyi Zhao and Zhenbang Sun},
booktitle={NeurIPS 2023 Workshop on Instruction Tuning and Instruction Following},
year={2023},
url={https://openreview.net/forum?id=HVduJbHSSO}
}

@article{hu2025prescribing,
  title={Prescribing the Right Remedy: Mitigating Hallucinations in Large Vision-Language Models via Targeted Instruction Tuning},
  author={Hu, Rui and Tu, Yahan and Wei, Shuyu and Lu, Dongyuan and Sang, Jitao},
  journal={Information Sciences},
  pages={122361},
  year={2025},
  publisher={Elsevier}
}

@inproceedings{
xiao2024efficient,
title={Efficient Streaming Language Models with Attention Sinks},
author={Guangxuan Xiao and Yuandong Tian and Beidi Chen and Song Han and Mike Lewis},
booktitle={The Twelfth International Conference on Learning Representations},
year={2024},
url={https://openreview.net/forum?id=NG7sS51zVF}
}

@article{zhang2025illusionbench,
  title={IllusionBench: A Large-scale and Comprehensive Benchmark for Visual Illusion Understanding in Vision-Language Models},
  author={Zhang, Yiming and Zhang, Zicheng and Wei, Xinyi and Liu, Xiaohong and Zhai, Guangtao and Min, Xiongkuo},
  journal={arXiv preprint arXiv:2501.00848},
  year={2025}
}

@inproceedings{
zhou2025mitigating,
title={Mitigating Modality Prior-Induced Hallucinations in Multimodal Large Language Models via Deciphering Attention Causality},
author={Guanyu Zhou and Yibo Yan and Xin Zou and Kun Wang and Aiwei Liu and Xuming Hu},
booktitle={The Thirteenth International Conference on Learning Representations},
year={2025},
url={https://openreview.net/forum?id=AV7OXVlAyi}
}

@inproceedings{yue2024mmmu,
  title={Mmmu: A massive multi-discipline multimodal understanding and reasoning benchmark for expert agi},
  author={Yue, Xiang and Ni, Yuansheng and Zhang, Kai and Zheng, Tianyu and Liu, Ruoqi and Zhang, Ge and Stevens, Samuel and Jiang, Dongfu and Ren, Weiming and Sun, Yuxuan and others},
  booktitle={Proceedings of the IEEE/CVF Conference on Computer Vision and Pattern Recognition},
  pages={9556--9567},
  year={2024}
}

@inproceedings{
parcalabescu2025do,
title={Do Vision \& Language Decoders use Images and Text equally? How Self-consistent are their Explanations?},
author={Letitia Parcalabescu and Anette Frank},
booktitle={The Thirteenth International Conference on Learning Representations},
year={2025},
url={https://openreview.net/forum?id=lCasyP21Bf}
}

@inproceedings{
barbero2025why,
title={Why do {LLM}s attend to the first token?},
author={Federico Barbero and Alvaro Arroyo and Xiangming Gu and Christos Perivolaropoulos and Petar Veli{\v{c}}kovi{\'c} and Razvan Pascanu and Michael M. Bronstein},
booktitle={Second Conference on Language Modeling},
year={2025},
url={https://openreview.net/forum?id=tu4dFUsW5z}
}

@inproceedings{yangunderstanding,
  title={Understanding and mitigating hallucination in large vision-language models via modular attribution and intervention},
  author={Yang, Tianyun and Li, Ziniu and Cao, Juan and Xu, Chang},
  booktitle={The Thirteenth International Conference on Learning Representations}, year = {2025}
}

@inproceedings{liu2024paying,
  title={Paying more attention to image: A training-free method for alleviating hallucination in lvlms},
  author={Liu, Shi and Zheng, Kecheng and Chen, Wei},
  booktitle={European Conference on Computer Vision},
  pages={125--140},
  year={2024},
  organization={Springer}
}

@inproceedings{
shi2025vision,
title={Vision Function Layer in Multimodal {LLM}s},
author={Cheng Shi and Yizhou Yu and Sibei Yang},
booktitle={The Thirty-ninth Annual Conference on Neural Information Processing Systems},
year={2025},
url={https://openreview.net/forum?id=nTc0LSqtqE}
}

@inproceedings{
kang2025see,
title={See What You Are Told: Visual Attention Sink in Large Multimodal Models},
author={Seil Kang and Jinyeong Kim and Junhyeok Kim and Seong Jae Hwang},
booktitle={The Thirteenth International Conference on Learning Representations},
year={2025},
url={https://openreview.net/forum?id=7uDI7w5RQA}
}

@inproceedings{
ghosh2025visual,
title={Visual Description Grounding Reduces Hallucinations and Boosts Reasoning in {LVLM}s},
author={Sreyan Ghosh and Chandra Kiran Reddy Evuru and Sonal Kumar and Utkarsh Tyagi and Oriol Nieto and Zeyu Jin and Dinesh Manocha},
booktitle={The Thirteenth International Conference on Learning Representations},
year={2025},
url={https://openreview.net/forum?id=3PRvlT8b1R}
}

@misc{liu2025seeingbelievingprobingdisconnect,
      title={Seeing but Not Believing: Probing the Disconnect Between Visual Attention and Answer Correctness in VLMs}, 
      author={Zhining Liu and Ziyi Chen and Hui Liu and Chen Luo and Xianfeng Tang and Suhang Wang and Joy Zeng and Zhenwei Dai and Zhan Shi and Tianxin Wei and Benoit Dumoulin and Hanghang Tong},
      year={2025},
      eprint={2510.17771},
      archivePrefix={arXiv},
      primaryClass={cs.AI},
      url={https://arxiv.org/abs/2510.17771}, 
}

@misc{sarkar2025mitigatinghallucinationsvisionlanguagemodels,
      title={Mitigating Hallucinations in Vision-Language Models through Image-Guided Head Suppression}, 
      author={Sreetama Sarkar and Yue Che and Alex Gavin and Peter A. Beerel and Souvik Kundu},
      year={2025},
      eprint={2505.16411},
      archivePrefix={arXiv},
      primaryClass={cs.CV},
      url={https://arxiv.org/abs/2505.16411}, 
}

@article{zheng2025modality,
  title={Modality Bias in LVLMs: Analyzing and Mitigating Object Hallucination via Attention Lens},
  author={Zheng, Haohan and Zhang, Zhenguo},
  journal={arXiv preprint arXiv:2508.02419},
  year={2025}
}

@inproceedings{sim2025can,
  title={Can vlms actually see and read? A survey on modality collapse in vision-language models},
  author={Sim, Mong Yuan and Zhang, Wei Emma and Dai, Xiang and Fang, Biaoyan},
  booktitle={Findings of the Association for Computational Linguistics: ACL 2025},
  pages={24452--24470},
  year={2025}
}

@article{team2024qwen2,
  title={Qwen2 technical report},
  author={Team, Qwen and others},
  journal={arXiv preprint arXiv:2407.10671},
  volume={2},
  number={3},
  year={2024}
}

@inproceedings{
gu2025when,
title={When Attention Sink Emerges in Language Models: An Empirical View},
author={Xiangming Gu and Tianyu Pang and Chao Du and Qian Liu and Fengzhuo Zhang and Cunxiao Du and Ye Wang and Min Lin},
booktitle={The Thirteenth International Conference on Learning Representations},
year={2025},
url={https://openreview.net/forum?id=78Nn4QJTEN}
}

@inproceedings{
liu2025reducing,
title={Reducing Hallucinations in Large Vision-Language Models via Latent Space Steering},
author={Sheng Liu and Haotian Ye and James Zou},
booktitle={The Thirteenth International Conference on Learning Representations},
year={2025},
url={https://openreview.net/forum?id=LBl7Hez0fF}
}

@inproceedings{gong-etal-2024-damro,
    title = "{DAMRO}: Dive into the Attention Mechanism of {LVLM} to Reduce Object Hallucination",
    author = "Gong, Xuan  and
      Ming, Tianshi  and
      Wang, Xinpeng  and
      Wei, Zhihua",
    editor = "Al-Onaizan, Yaser  and
      Bansal, Mohit  and
      Chen, Yun-Nung",
    booktitle = "Proceedings of the 2024 Conference on Empirical Methods in Natural Language Processing",
    month = nov,
    year = "2024",
    address = "Miami, Florida, USA",
    publisher = "Association for Computational Linguistics",
    url = "https://aclanthology.org/2024.emnlp-main.439/",
    doi = "10.18653/v1/2024.emnlp-main.439",
    pages = "7696--7712"
}

@inproceedings{liu2023visual,
title={Visual Instruction Tuning},
author={Haotian Liu and Chunyuan Li and Qingyang Wu and Yong Jae Lee},
booktitle={Thirty-seventh Conference on Neural Information Processing Systems},
year={2023},
url={https://openreview.net/forum?id=w0H2xGHlkw}
}

@misc{shu2025largevisionlanguagemodelalignment,
      title={Large Vision-Language Model Alignment and Misalignment: A Survey Through the Lens of Explainability}, 
      author={Dong Shu and Haiyan Zhao and Jingyu Hu and Weiru Liu and Ali Payani and Lu Cheng and Mengnan Du},
      year={2025},
      eprint={2501.01346},
      archivePrefix={arXiv},
      primaryClass={cs.CV},
      url={https://arxiv.org/abs/2501.01346}, 
}

@inproceedings{
neo2025towards,
title={Towards Interpreting Visual Information Processing in Vision-Language Models},
author={Clement Neo and Luke Ong and Philip Torr and Mor Geva and David Krueger and Fazl Barez},
booktitle={The Thirteenth International Conference on Learning Representations},
year={2025},
url={https://openreview.net/forum?id=chanJGoa7f}
}

@inproceedings{
chen2024multiobject,
title={Multi-Object Hallucination in Vision Language Models},
author={Xuweiyi Chen and Ziqiao Ma and Xuejun Zhang and Sihan Xu and Shengyi Qian and Jianing Yang and David Fouhey and Joyce Chai},
booktitle={The Thirty-eighth Annual Conference on Neural Information Processing Systems},
year={2024},
url={https://openreview.net/forum?id=KNrwaFEi1u}
}

@inproceedings{
li2023evaluating,
title={Evaluating Object Hallucination in Large Vision-Language Models},
author={Yifan Li and Yifan Du and Kun Zhou and Jinpeng Wang and Xin Zhao and Ji-Rong Wen},
booktitle={The 2023 Conference on Empirical Methods in Natural Language Processing},
year={2023},
url={https://openreview.net/forum?id=xozJw0kZXF}
}

@inproceedings{yebin2024beaf,
  title     = {BEAF: Observing BEfore-AFter Changes to Evaluate Hallucination in Vision-language Models},
  author    = {Moon, Ye-Bin and Nam, Hyeon-Woo and Choi, Wonseok and Oh, Tae-Hyun},
  booktitle = {European Conference on Computer Vision (ECCV)},
  year      = {2024},
}

@inproceedings{
wang2025mllm,
title={{MLLM} can see? Dynamic Correction Decoding for Hallucination Mitigation},
author={Chenxi Wang and Xiang Chen and Ningyu Zhang and Bozhong Tian and Haoming Xu and Shumin Deng and Huajun Chen},
booktitle={The Thirteenth International Conference on Learning Representations},
year={2025},
url={https://openreview.net/forum?id=4z3IguA4Zg}
}

@article{DBLP:journals/corr/abs-2404-18930,
  publtype={informal},
  author={Zechen Bai and Pichao Wang and Tianjun Xiao and Tong He and Zongbo Han and Zheng Zhang and Mike Zheng Shou},
  title={Hallucination of Multimodal Large Language Models: A Survey},
  year={2024},
  cdate={1704067200000},
  journal={CoRR},
  volume={abs/2404.18930},
  url={https://doi.org/10.48550/arXiv.2404.18930}
}

@inproceedings{huang2024opera,
  title={Opera: Alleviating hallucination in multi-modal large language models via over-trust penalty and retrospection-allocation},
  author={Huang, Qidong and Dong, Xiaoyi and Zhang, Pan and Wang, Bin and He, Conghui and Wang, Jiaqi and Lin, Dahua and Zhang, Weiming and Yu, Nenghai},
  booktitle={Proceedings of the IEEE/CVF Conference on Computer Vision and Pattern Recognition},
  pages={13418--13427},
  year={2024}
}

@inproceedings{chen2024image,
  title={An image is worth 1/2 tokens after layer 2: Plug-and-play inference acceleration for large vision-language models},
  author={Chen, Liang and Zhao, Haozhe and Liu, Tianyu and Bai, Shuai and Lin, Junyang and Zhou, Chang and Chang, Baobao},
  booktitle={European Conference on Computer Vision},
  pages={19--35},
  year={2024},
  organization={Springer}
}

@inproceedings{lin2014microsoft,
  title={Microsoft coco: Common objects in context},
  author={Lin, Tsung-Yi and Maire, Michael and Belongie, Serge and Hays, James and Perona, Pietro and Ramanan, Deva and Doll{\'a}r, Piotr and Zitnick, C Lawrence},
  booktitle={Computer vision--ECCV 2014: 13th European conference, zurich, Switzerland, September 6-12, 2014, proceedings, part v 13},
  pages={740--755},
  year={2014},
  organization={Springer}
}

@InProceedings{Guan_2024_CVPR,
    author    = {Guan, Tianrui and Liu, Fuxiao and Wu, Xiyang and Xian, Ruiqi and Li, Zongxia and Liu, Xiaoyu and Wang, Xijun and Chen, Lichang and Huang, Furong and Yacoob, Yaser and Manocha, Dinesh and Zhou, Tianyi},
    title     = {HallusionBench: An Advanced Diagnostic Suite for Entangled Language Hallucination and Visual Illusion in Large Vision-Language Models},
    booktitle = {Proceedings of the IEEE/CVF Conference on Computer Vision and Pattern Recognition (CVPR)},
    month     = {June},
    year      = {2024},
    pages     = {14375-14385}
}

@inproceedings{zhu2025ibd,
  title={Ibd: Alleviating hallucinations in large vision-language models via image-biased decoding},
  author={Zhu, Lanyun and Ji, Deyi and Chen, Tianrun and Xu, Peng and Ye, Jieping and Liu, Jun},
  booktitle={Proceedings of the Computer Vision and Pattern Recognition Conference},
  pages={1624--1633},
  year={2025}
}

@inproceedings{bi2025unveiling,
  title={Unveiling visual perception in language models: An attention head analysis approach},
  author={Bi, Jing and Guo, Junjia and Tang, Yunlong and Wen, Lianggong Bruce and Liu, Zhang and Wang, Bingjie and Xu, Chenliang},
  booktitle={Proceedings of the Computer Vision and Pattern Recognition Conference},
  pages={4135--4144},
  year={2025}
}

@inproceedings{fu2025mme,
  title={Mme: A comprehensive evaluation benchmark for multimodal large language models},
  author={Fu, Chaoyou and Chen, Peixian and Shen, Yunhang and Qin, Yulei and Zhang, Mengdan and Lin, Xu and Yang, Jinrui and Zheng, Xiawu and Li, Ke and Sun, Xing and others},
  booktitle={The Thirty-ninth Annual Conference on Neural Information Processing Systems Datasets and Benchmarks Track},
  year={2025}
}

@article{hsieh2023sugarcrepe,
  title={Sugarcrepe: Fixing hackable benchmarks for vision-language compositionality},
  author={Hsieh, Cheng-Yu and Zhang, Jieyu and Ma, Zixian and Kembhavi, Aniruddha and Krishna, Ranjay},
  journal={Advances in neural information processing systems},
  volume={36},
  pages={31096--31116},
  year={2023}
}

@article{li2024naturalbench,
  title={Naturalbench: Evaluating vision-language models on natural adversarial samples},
  author={Li, Baiqi and Lin, Zhiqiu and Peng, Wenxuan and Nyandwi, Jean de Dieu and Jiang, Daniel and Ma, Zixian and Khanuja, Simran and Krishna, Ranjay and Neubig, Graham and Ramanan, Deva},
  journal={Advances in Neural Information Processing Systems},
  volume={37},
  pages={17044--17068},
  year={2024}
}

@article{liu2024survey,
  title={A Survey on Hallucination in Large Vision-Language Models},
  author={Liu, Hanchao and Xue, Wenyuan and Chen, Yifei and Chen, Dapeng and Zhao, Xiutian and Wang, Ke and Hou, Liping and Li, Rongjun and Peng, Wei},
  journal={CoRR},
  year={2024}
}

@misc{
liang2026not,
title={{Not Errors but Guardians: Understanding Sink Tokens in Multimodal LLMs}},
author={Haiji Liang and Yujun Cai},
year={2026},
url={https://openreview.net/forum?id=EpoJKtVxNt},
note={{OpenReview}}
}

\appendix

\section{Appendix}
\label{sec:appendix}

%\raggedbottom
\subsection{Baseline methods used in Q4 proportional redistribution}
\label{app:baseline_prop}

As stated in Section \ref{sec:experiment}, we include two methods that boost image attention and one that suppresses text attention:

Image-boosting:
\begin{itemize}
    \item \textbf{PAI} \citep{liu2024paying} scales image weights by 1.5 \textit{before} softmax, then ensures the output logits are sufficiently distinct from those of a unimodal model without access to image inputs. It targets all decoder heads by default. We use a reimplementation of the original algorithm, with the recommended settings of $\alpha = 0.5$ and $\gamma = 1.1$.
    \item \textbf{$\text{Image}\times2.0$} is an ad-hoc intervention on the decoder attention heads identified by \citet{yangunderstanding} to contribute most to object hallucination in the MSCOCO \citep{lin2014microsoft} image captioning task: their `MSCOCO hallucination heads'. It simply scales image attention on each by 2.0.
\end{itemize}

Text-suppressing:
\begin{itemize}
    \item \textbf{AD-HH} \citep{yangunderstanding} similarly targets \citeposs{yangunderstanding} MSCOCO hallucination heads. Following the recommended settings, text attention is zeroed out on a head if it exceeds 40\% of the total weights post softmax.
\end{itemize}

\FloatBarrier

\subsection{The yes-bias is a late-layer phenomenon}
\label{app:late_layer}
This section provides empirical justification for our focus on Q4. We first rule out the presence of any substantial attention imbalances at the global (i.e.~whole-model) level and subsequently in Q1-3 (i.e.~layers 1 to 24) of LLaVA-1.5 7B's text decoder.

\subsubsection{Global redistributions}
\label{app:global}

To assess whether the yes-bias arises from global attention imbalances as a cause of the yes-bias, we perform both pairwise and proportional redistributions of attention weights at the global level, intervening simultaneously across all decoder heads. Unlike the 100\% redistributions employed in earlier experiments, we adopt graduated transfers that reallocate 10\%, 20\% and 30\% of the source modality's attention weights respectively. This is because fine-tuning showed that transferring more than 30\% of weights from the source modality across all decoder heads tended to lead to a collapse of model functionality. 

\begin{figure*}[t]
  \centering

  \begin{subfigure}{0.48\linewidth}
    \includegraphics[width=\linewidth]{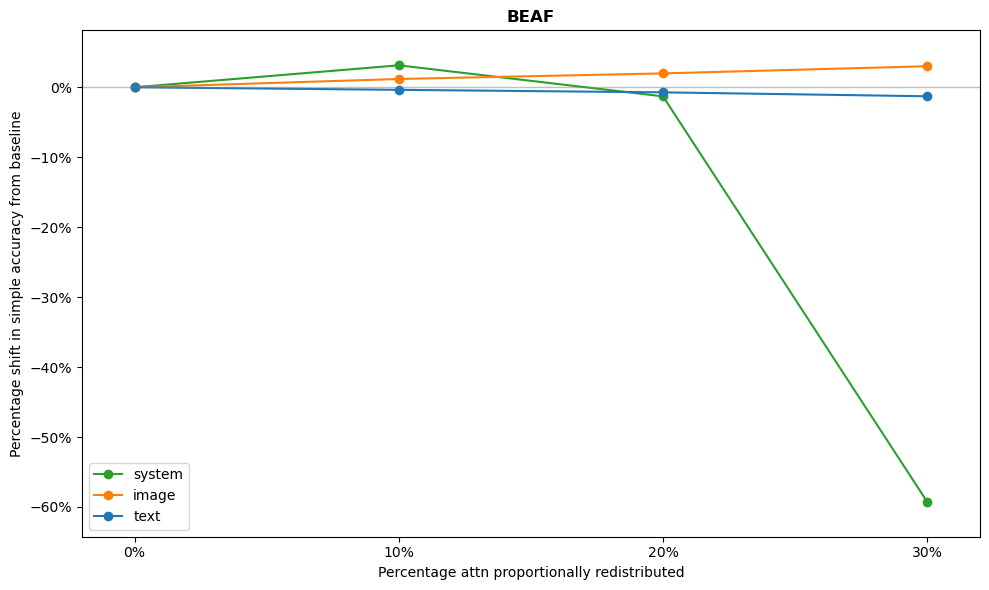}
  \end{subfigure}\hfill
  \begin{subfigure}{0.48\linewidth}
    \includegraphics[width=\linewidth]{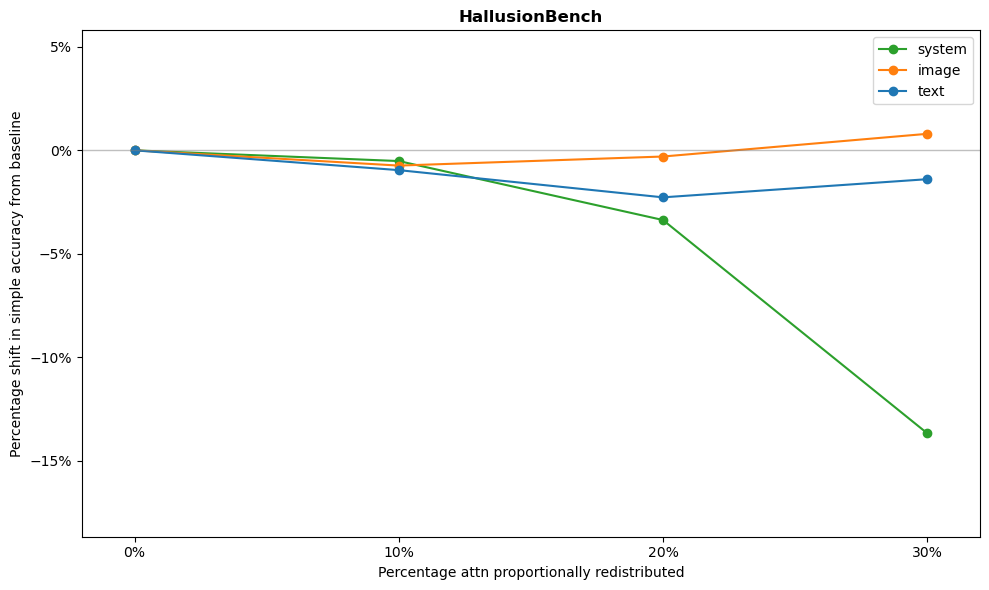}
  \end{subfigure}

  \vspace{0.5em}

  \begin{subfigure}{0.48\linewidth}
    \includegraphics[width=\linewidth]{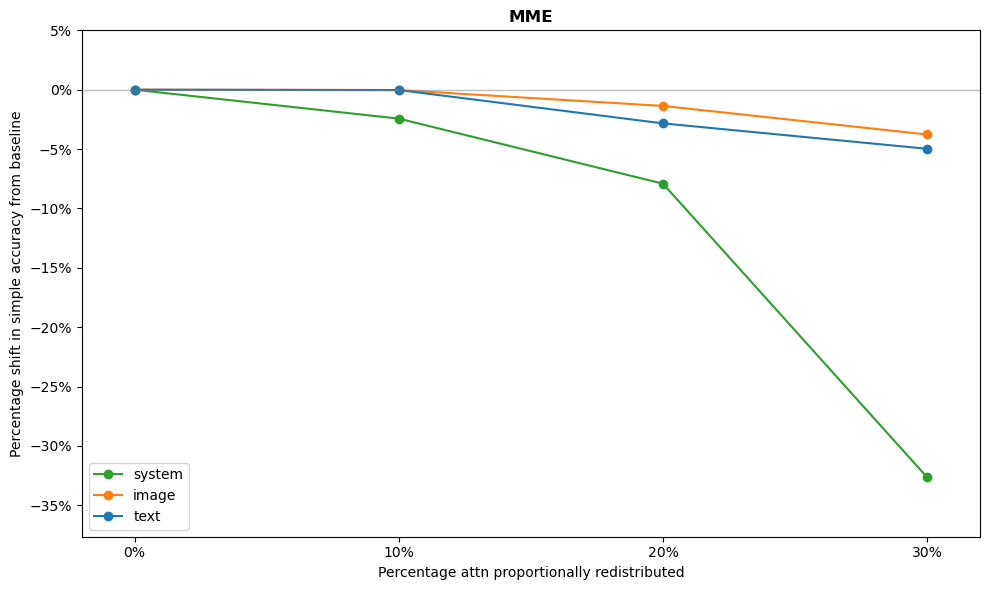}
  \end{subfigure}\hfill
  \begin{subfigure}{0.48\linewidth}
    \includegraphics[width=\linewidth]{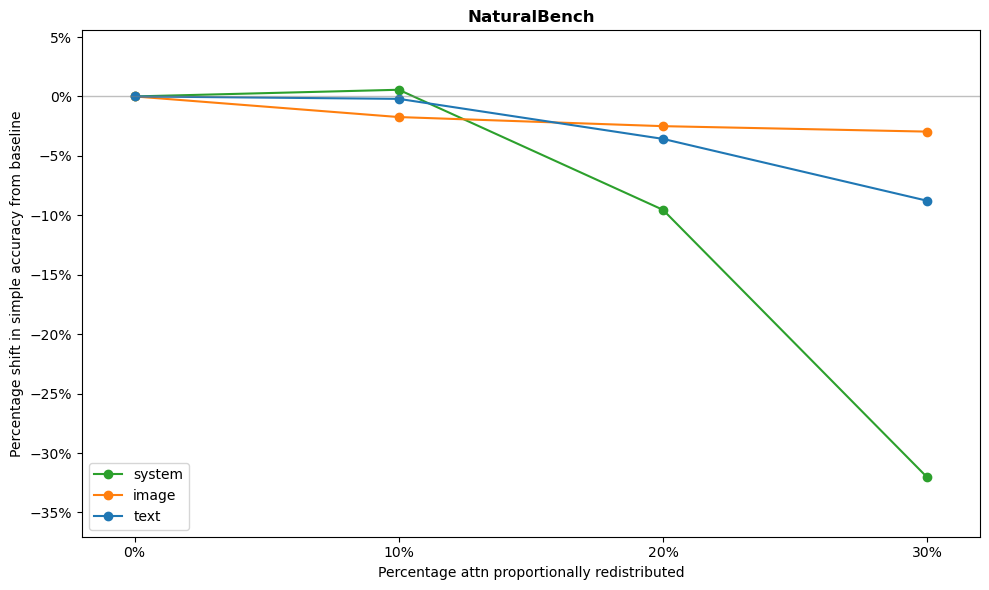}
  \end{subfigure}

  \vspace{0.5em}

  \begin{subfigure}{0.48\linewidth}
    \includegraphics[width=\linewidth]{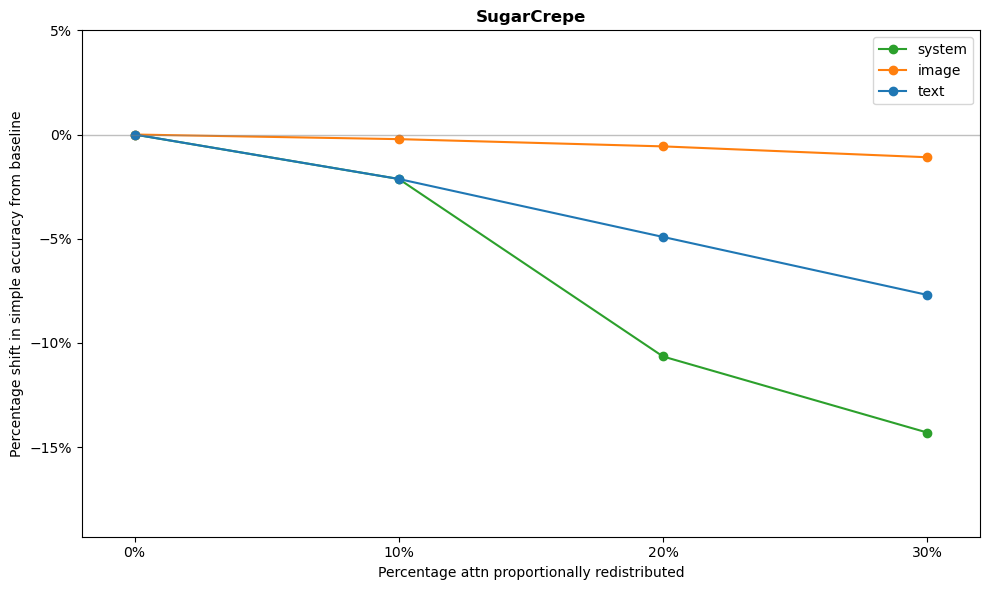}
  \end{subfigure}\hfill
  \begin{subfigure}{0.48\linewidth}
    \includegraphics[width=\linewidth]{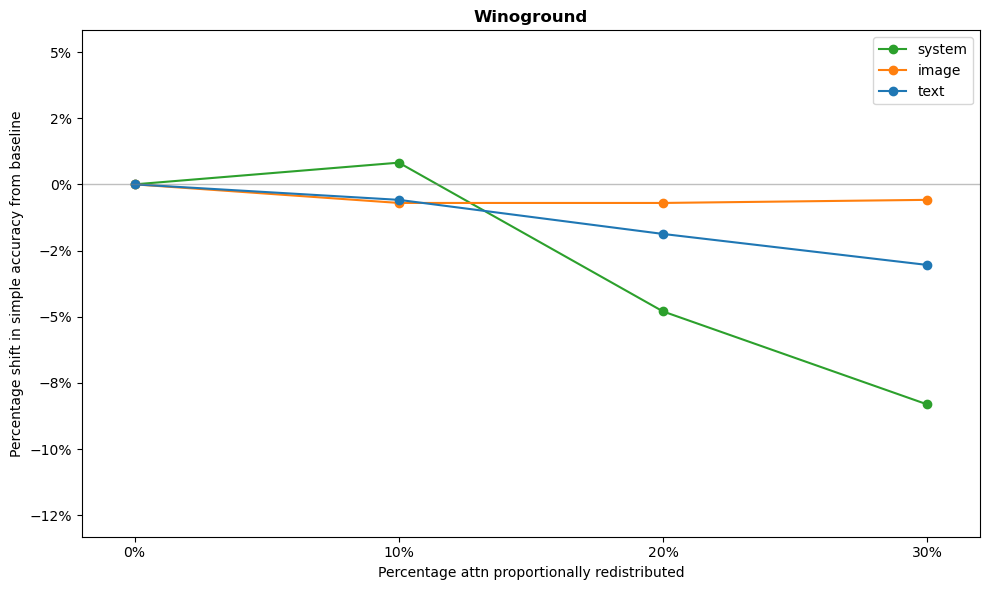}
  \end{subfigure}

  \caption{Percentage shift in simple accuracy against percentage of attention weights from each source modality proportionally redistributed to the remaining modalities \textit{across all decoder heads} (i.e.~globally), for all six paired-prompt benchmarks in our evaluation suite. We observe that the majority of these interventions are harmful and gains are not consistent across benchmarks. Despite not exhibiting the same types of gains elsewhere, MME patterns with the rest of the benchmarks as far as the harmfulness of global interventions is concerned.}
  \label{fig:global_all_benchmarks}
\end{figure*}

\begin{figure*}[t!]
  \centering

  \begin{subfigure}{0.55\linewidth}
    \includegraphics[width=\linewidth]{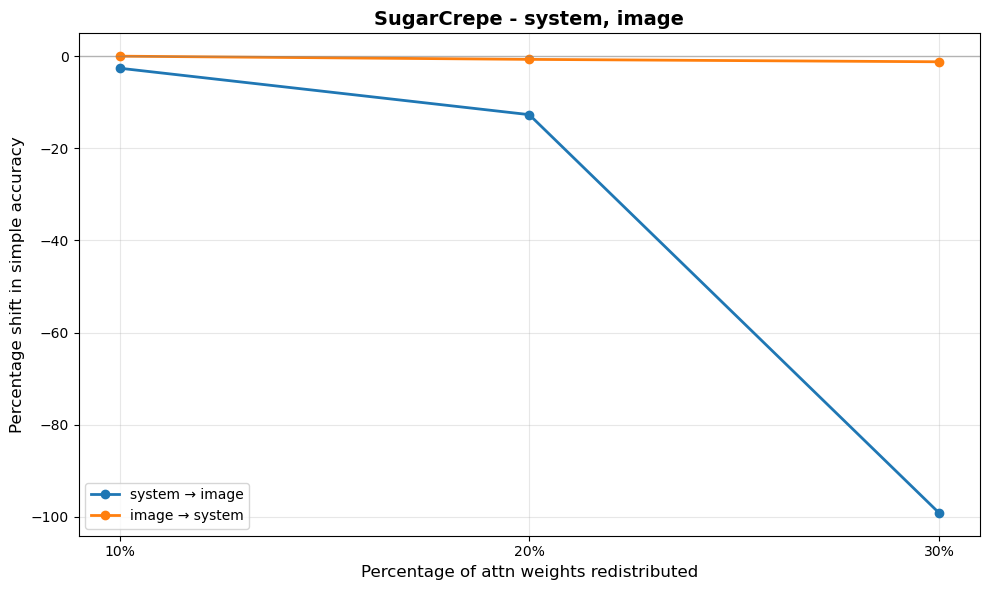}
  \end{subfigure}

  \vspace{0.5em}

  \begin{subfigure}{0.55\linewidth}
    \includegraphics[width=\linewidth]{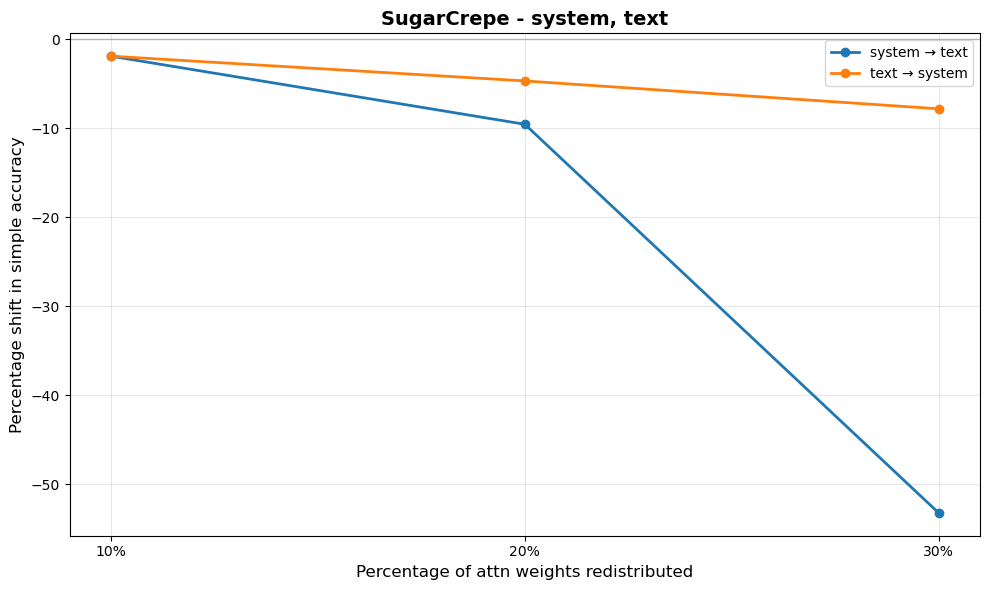}
  \end{subfigure}

  \vspace{0.5em}

  \begin{subfigure}{0.55\linewidth}
    \includegraphics[width=\linewidth]{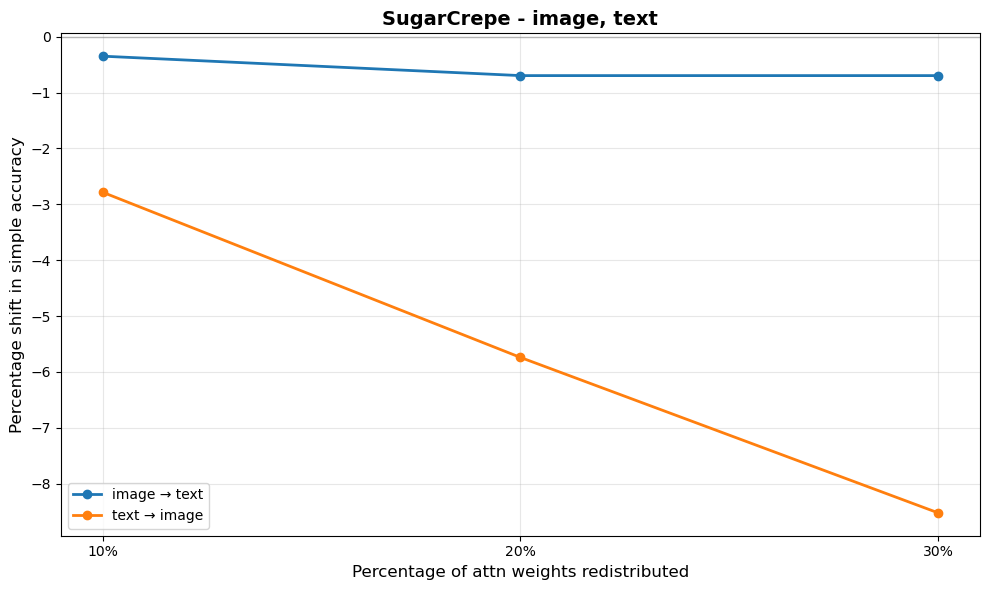}
  \end{subfigure}

  \caption{Global (i.e.~applied across all decoder heads) graduated pairwise redistributions of attention weights between all possible combinations of source and recipient modalities in SugarCrepe show whole-model interventions to mostly harm performance. These soft-touch interventions redistribute 10\%, 20\% and 30\% of weights from source to recipient at a time. The same trend is reflected across all six benchmarks studied here, but results are not shown owing to space constraints.}
  \label{fig:global_sc_prop}
\end{figure*}

Overall, these global interventions predominantly degrade performance. In the few cases where performance does not deteriorate, the observed improvements are marginal and substantially smaller than those achieved through the Q4 interventions introduced in Sections \ref{sec:framework} and \ref{sec:experiment}. This suggests that global attention imbalances are neither robust nor the primary driver of the yes-bias, thereby supporting the localised intervention strategy adopted in this work. Where proportional redistributions of attention weights are concerned, Figure \ref{fig:global_all_benchmarks} shows these to be mostly harmful regardless of the source and recipient modalities involved and not lead to consistent gains. 
Figure \ref{fig:global_sc_prop} tells a similar story for proportional redistributions in the case of the SugarCrepe benchmark. The remaining plots and metrics beyond simple accuracy are not shown for space considerations, >100 interventions having been performed here. However, the observed trends are consistent across our benchmarks and evaluation metrics, also extending to the pairwise setting.

These results justify our local approach to interventions, emphasising redistributions performed at the quarter level rather than across all decoder heads.

\subsubsection{Per-quarter redistributions}
\label{app:quarter}
\begin{figure*}[t]
  \centering

  \begin{subfigure}{0.48\linewidth}
    \includegraphics[width=\linewidth]{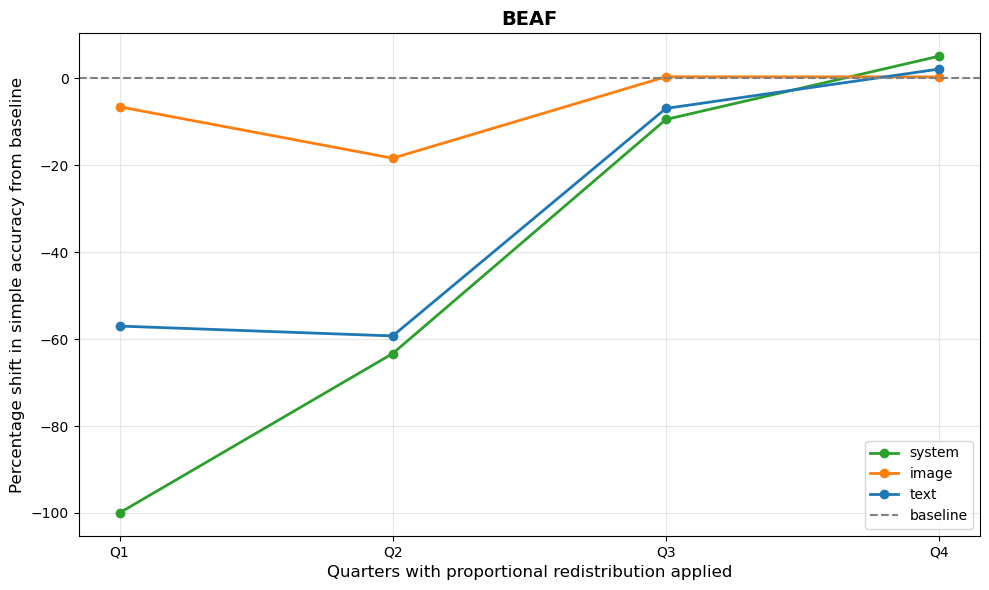}
  \end{subfigure}\hfill
  \begin{subfigure}{0.48\linewidth}
    \includegraphics[width=\linewidth]{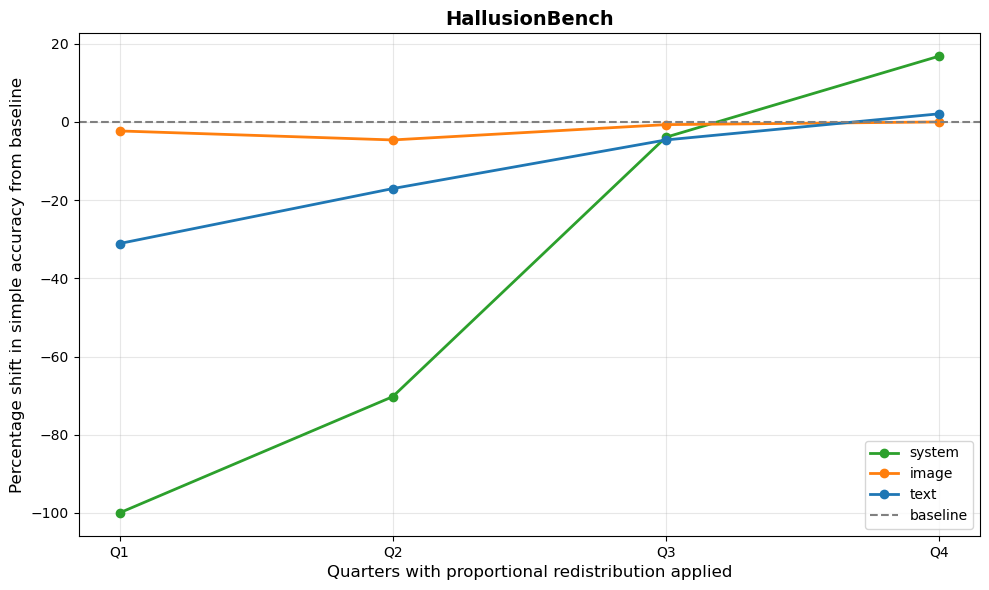}
  \end{subfigure}

  \vspace{0.5em}

  \begin{subfigure}{0.48\linewidth}
    \includegraphics[width=\linewidth]{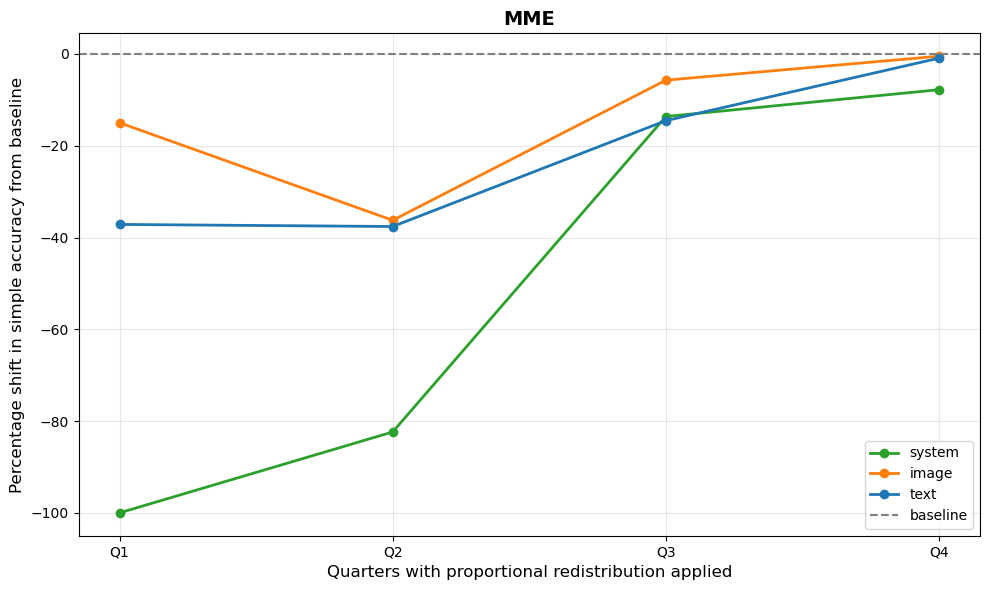}
  \end{subfigure}\hfill
  \begin{subfigure}{0.48\linewidth}
    \includegraphics[width=\linewidth]{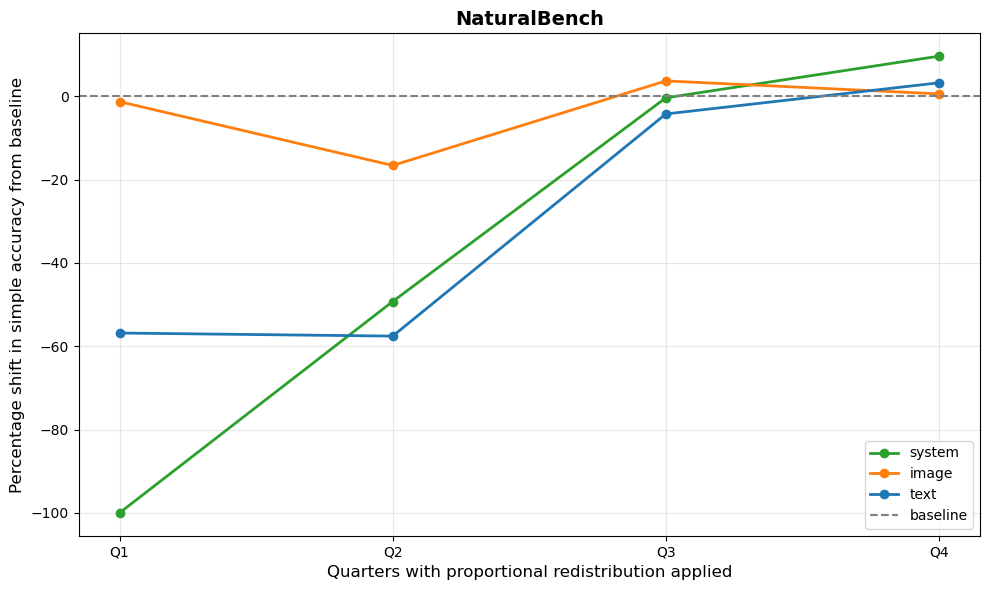}
  \end{subfigure}

  \vspace{0.5em}

  \begin{subfigure}{0.48\linewidth}
    \includegraphics[width=\linewidth]{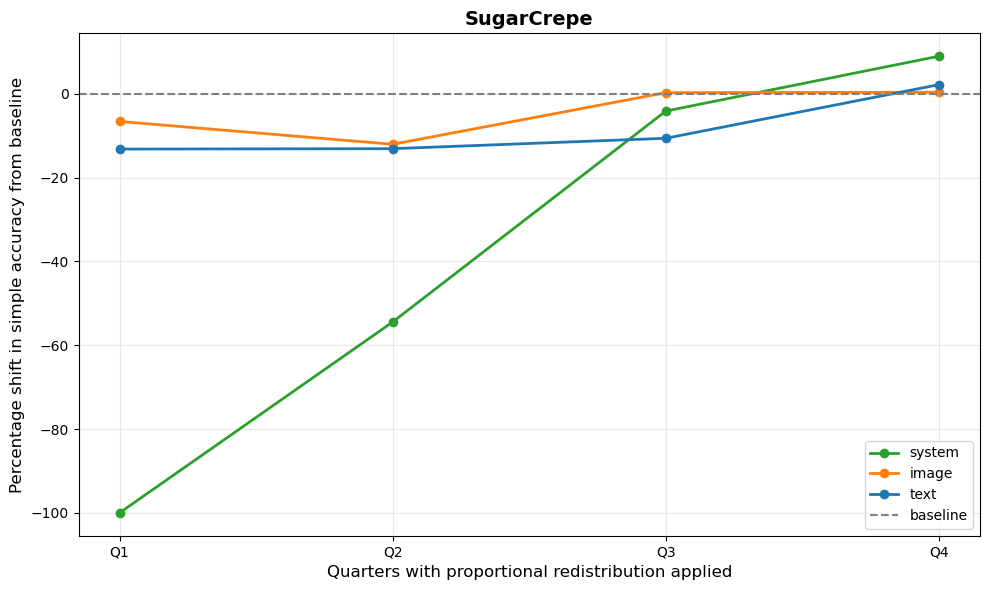}
  \end{subfigure}\hfill
  \begin{subfigure}{0.48\linewidth}
    \includegraphics[width=\linewidth]{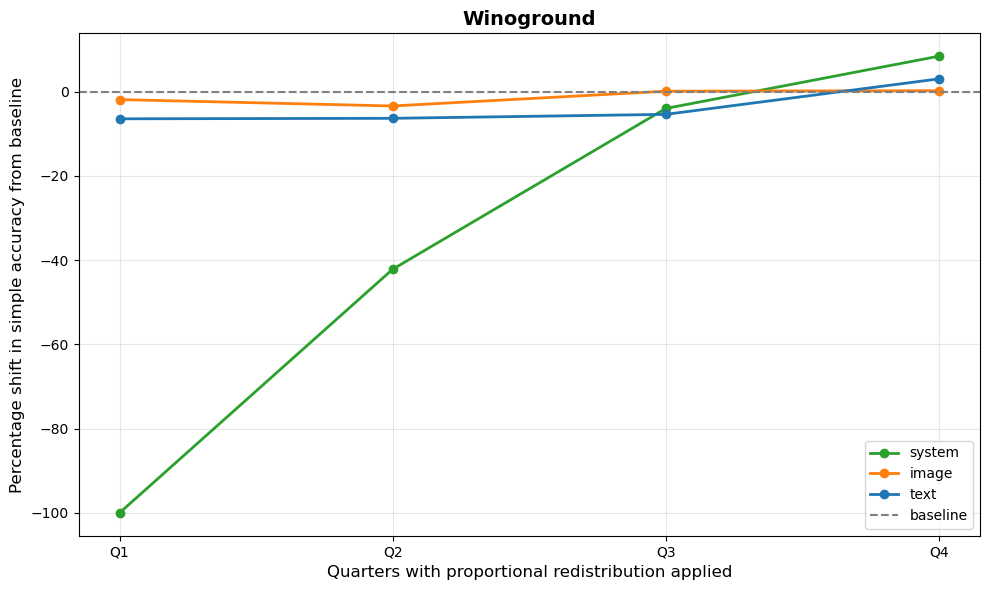}
  \end{subfigure}

  \caption{Percentage shift in simple accuracy against percentage of attention weights from each source modality proportionally redistributed to the remaining modalities \textit{per quarter}, for all six paired-prompt benchmarks in our evaluation suite. These plots demonstrate that for five of our six benchmarks, proportional redistribution of system attention in Q4 yields the largest gains. MME is the sole exception to this pattern.}
  \label{fig:quarter_all_benchmarks}
\end{figure*}

To verify that Q4 is indeed where the largest attention imbalances are found, we carried out pairwise and proportional redistribution of attention weights between all possible combinations of modalities in Q1-3 in turn. As in prior quarter-level experiments, these interventions involve redistributing 100\% of the attention weights from the source modality.
In terms of proportional redistribution, Figure \ref{fig:quarter_all_benchmarks} shows that across all six benchmarks, interventions in Q1 and Q2 are consistently detrimental, regardless of the source or target modality. These trends hold even for MME, the one benchmark that Q4 system redistribution failed to yield enhanced performance for. Interventions in Q3 similarly fail to yield consistent improvements.

Accuracy gains are almost exclusively concentrated in Q4 (layers 25-32), implicating the system modality most of all but to a much smaller extent also the textual one. This indicates that the earlier portions of the decoder do not contain stable reservoirs of modality-specific redundant attention that can be reallocated for performance gains comparable to those achieved in Q4. 

Unexpectedly by our assumption that there is insufficient image attention in Q4, redistributing image weights in that quarter does not make a substantial difference to performance. At first glance, this appears to confirm observations by \citet{shu2025largevisionlanguagemodelalignment} and \citet{bi2025unveiling} that image attention in the later decoder layers may be largely redundant. What is problematic about such an account is that boosting image attention by redistributing system and text attention in Q4 benefits performance, suggesting that there may be more subtle deficits in image attention there after all.

\begin{figure*}[t!]
  \centering

  \begin{subfigure}{0.55\linewidth}
    \includegraphics[width=\linewidth]{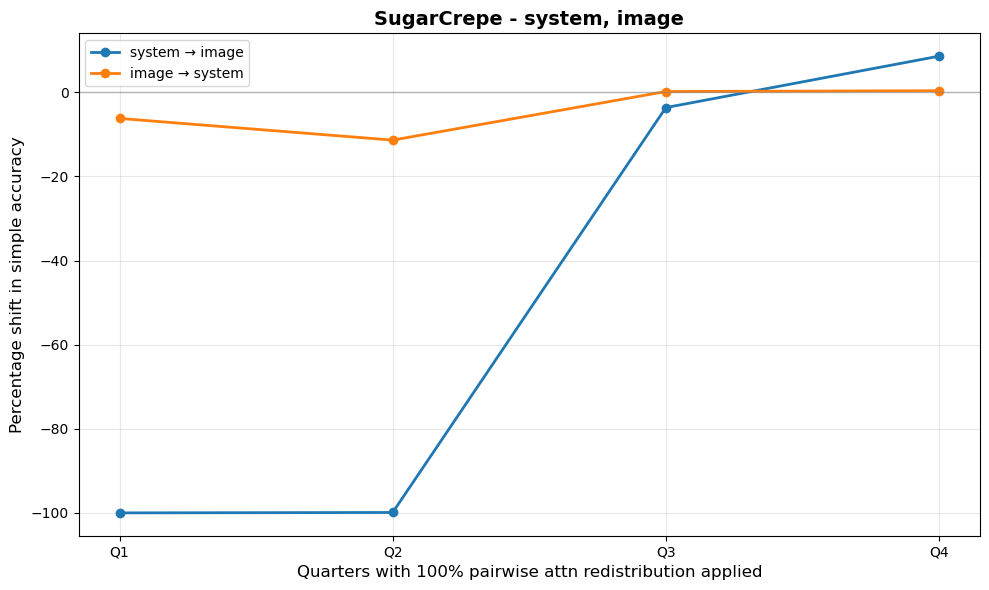}
  \end{subfigure}

  \vspace{0.5em}

  \begin{subfigure}{0.55\linewidth}
    \includegraphics[width=\linewidth]{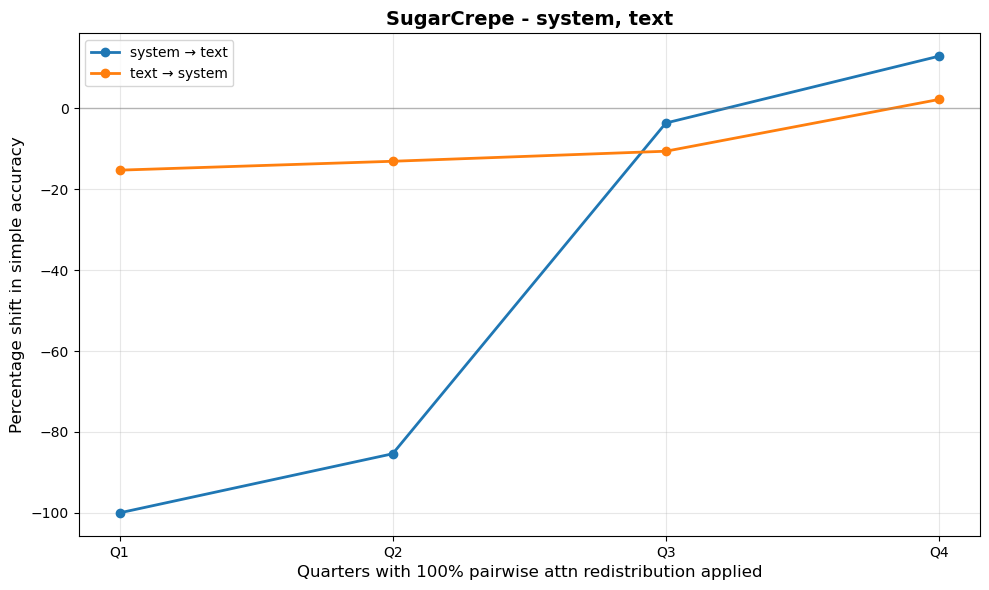}
  \end{subfigure}

  \vspace{0.5em}

  \begin{subfigure}{0.55\linewidth}
    \includegraphics[width=\linewidth]{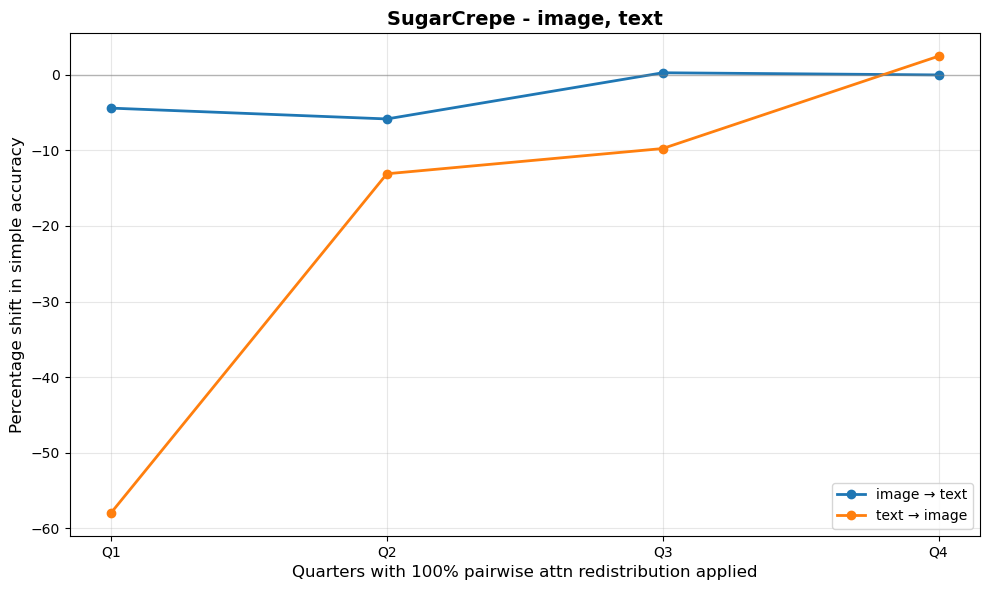}
  \end{subfigure}

  \caption{Per-quarter 100\% pairwise redistributions of attention weights between all possible combinations of source and recipient modalities in SugarCrepe show whole-model interventions to mostly harm performance. The same trend is reflected across all six benchmarks studied here, but results are not shown owing to space constraints.}
  \label{fig:quarter_sc}
\end{figure*}

Now turning to pairwise redistributions of weights, Figure \ref{fig:quarter_sc} shows that gains through these interventions are similarly almost exclusively concentrated in Q4. While only results for SugarCrepe are reported, the same trends obtain across our other paired-prompt benchmarks. 

\begin{table*}[t!]
  \centering
  \small
  \renewcommand{\arraystretch}{1.8}
  \begin{tabular}{m{0.12\textwidth} m{0.22\textwidth} m{0.30\textwidth} m{0.30\textwidth}}
    \hline
    \textbf{Category} & \textbf{Image} & \textbf{Positive prompt} & \textbf{Negative prompt} \\
    \hline
    Artwork
    &
    \centering
    \par\vspace{5pt}
    \includegraphics[width=0.9\linewidth,keepaspectratio]{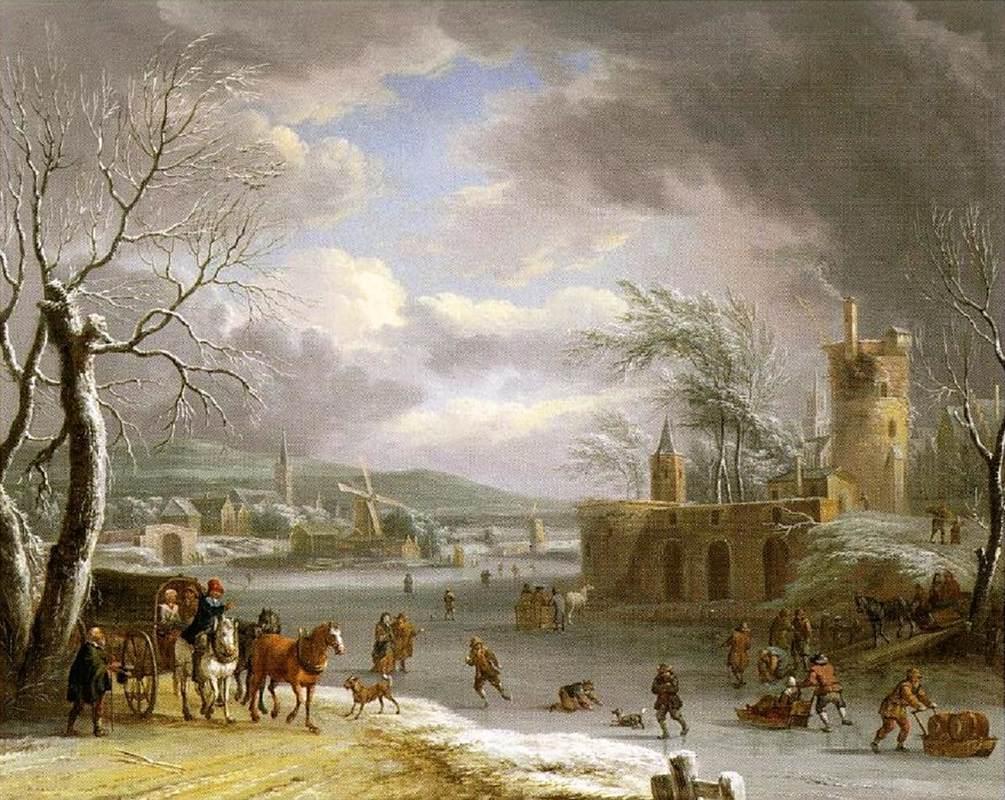}
    \par\vspace{2pt}
    &
    Does this artwork exist in the form of painting? Please answer yes or no.
    &
    Does this artwork exist in the form of sculpture? Please answer yes or no.
    \\
    \hline
    Landmark
    &
    \centering
    \par\vspace{5pt}
    \includegraphics[width=0.9\linewidth,keepaspectratio]{latex/mme_landmark_10b2b8741f283a3c.jpg}
    \par\vspace{2pt}
    &
    Is this a photo of De Bataaf, Winterswijk? Please answer yes or no.
    &
    Is this a photo of Porte Guillaume-Lion? Please answer yes or no.
    \\
    \hline
    Posters
    &
    \centering
    \par\vspace{5pt}
    \includegraphics[width=0.9\linewidth,keepaspectratio]{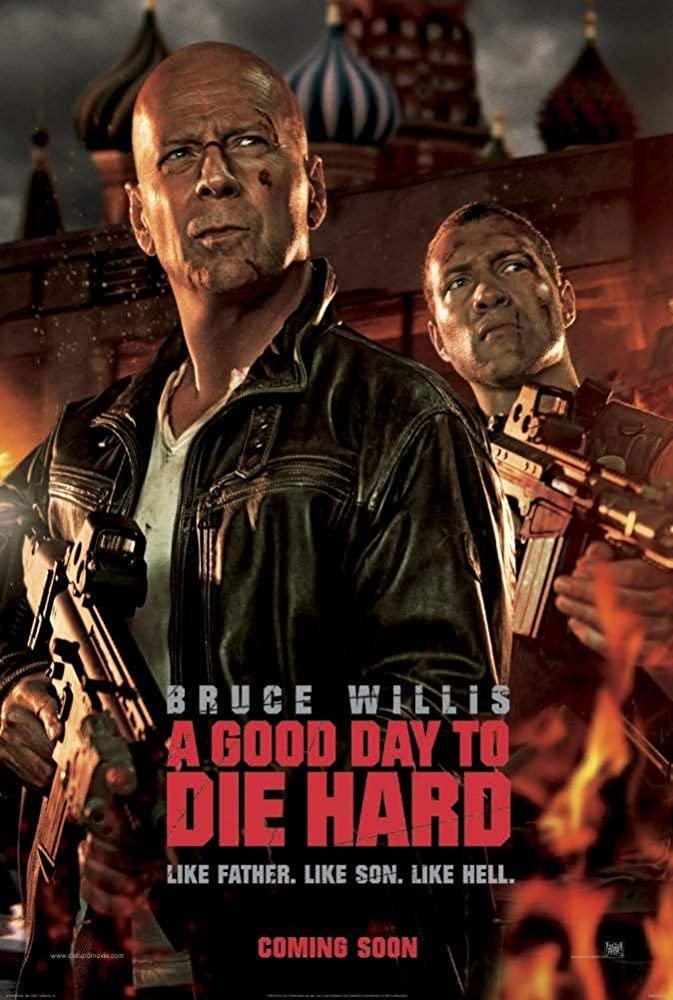}
    \par\vspace{2pt}
    &
    Is this movie originated from the country or region of usa? Please answer yes or no.
    &
    Is this movie originated from the country or region of uk? Please answer yes or no.
    \\
    \hline
    Scene
    &
    \centering
    \par\vspace{5pt}
    \includegraphics[width=0.9\linewidth,keepaspectratio]{latex/mme_scene_Places365_val_00000156.jpg}
    \par\vspace{2pt}
    &
    Does this image describe a place of stable? Please answer yes or no.
    &
    Does this image describe a place of junkyard? Please answer yes or no.
    \\
    \hline
  \end{tabular}
  \caption{\label{tab:mme_examples}
    Example prompts from MME, which may permit reliance on high-level, global cross-modal representations. Positive prompts refer to those with ground truth `yes' and negative to those with ground truth `no'.
  }
\end{table*}

\begin{table*}[t]
  \centering
  \small
  \renewcommand{\arraystretch}{1.8}
  \begin{tabular}{m{0.12\textwidth} m{0.20\textwidth} m{0.32\textwidth} m{0.32\textwidth}}
    \hline
    \textbf{Category} & \textbf{Image} & \textbf{Positive prompts} & \textbf{Negative prompt} \\
    \hline
    Adversarial
    &
    \multirow{3}{=}[-1em]{\centering
    \includegraphics[width=0.9\linewidth,keepaspectratio]{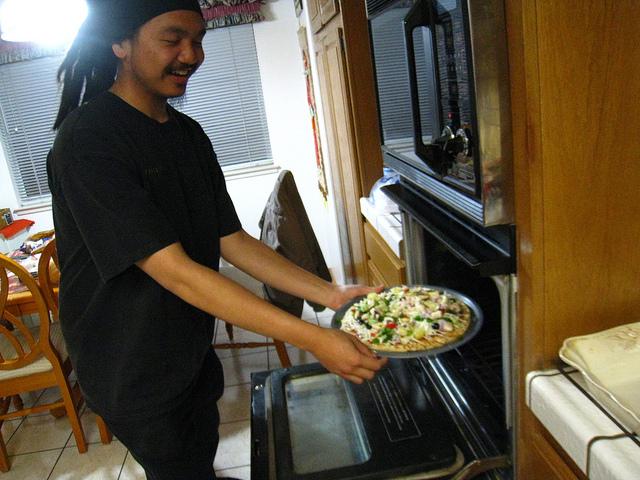}
    \par\vspace{2pt}}
    &
    \multirow{3}{=}[-1.5em]{Is there a \{chair/ person/ dining table\} in the image? Answer only `yes' or `no'.}
    &
    Is there a couch in the image? Answer only `yes' or `no'.
    \\
    \cline{1-1}\cline{4-4}
    Popular
    &
    &
    &
    Is there a car in the image? Answer only `yes' or `no'.
    \\
    \cline{1-1}\cline{4-4}
    Random
    &
    &
    &
    Is there an airplane in the image? Answer only `yes' or `no'.
    \\
    \hline
  \end{tabular}
  \caption{\label{tab:pope_examples}
    Example prompts from POPE demonstrating different sampling strategies.
  }
\end{table*}

\subsection{Example prompts}
\label{app:examples}
In this appendix, we present example prompts from MME (Table \ref{tab:mme_examples}), the three splits of POPE (Table \ref{tab:pope_examples}) and the remaining paired-prompt benchmarks (Table \ref{tab:paired_prompt_examples}) to illustrate the contrast between MME and the adversarial POPE split on the one hand and the rest of the benchmarks on the other, as described in Section \ref{sec:error_analysis}.

\subsection{Yes/no vs other responses}
\label{app:other_responses}
Here, we put forth additional evidence for a causal relationship between the yes-bias and redundant late-layer system attention. This is important in practical terms because it suggests that the yes-bias can be mitigated by targeting one clearly defined underlying factor.

To this end, we adduce two further sets of prompts to demonstrate that Q4 system redistribution most strongly impacts on yes/no responses. Other types of response, on the other hand, do not exhibit the same changes under intervention.

\noindent\textbf{\textit{NaturalBench (A/B prompts)}}.~The first set of prompts is the A/B subset of the NaturalBench dataset. NaturalBench comprises two types of binary-choice prompt, the first yes/no --- covered in the main body (see Table \ref{tab:benchmarks}) --- and the second A/B. 

Despite both involving binary choice, Q4 system redistribution only exerts a markedly beneficial effect on the former (+60.51\% paired accuracy, Table \ref{tab:nb_yn_ab}). For A/B prompts, on the other hand, the intervention only marginally improves performance (+1.01\%). This strongly indicates that the mechanism described in this study, redundant late-layer system attention, is highly specific to yes-bias hallucination.

\newcolumntype{C}[1]{>{\centering\arraybackslash}p{#1}}

\begin{table*}[t]
  \centering
  \footnotesize
  \renewcommand{\arraystretch}{1.05}
  \begin{tabular}{p{0.14\textwidth} C{0.16\textwidth} C{0.16\textwidth} | C{0.16\textwidth} C{0.16\textwidth}}
    \toprule
    \multirow{2}{*}{\textbf{NaturalBench}} & \multicolumn{2}{c}{\textbf{Yes/no prompts}} & \multicolumn{2}{c}{\textbf{A/B prompts}} \\
    \cmidrule{2-5}
    & \textit{Q4 system redistr (prop)} & \textit{No-intervention baseline} & \textit{Q4 system redistr (prop)} & \textit{No-intervention baseline} \\
    \midrule
    Simple acc$^{\Uparrow}$ & 66.02 (\textbf{+9.63}) & 60.22 & 69.33 (+0.57) & 68.94 \\
    Paired acc$^{\Uparrow}$ & 34.59 (\textbf{+60.51}) & 21.55 & 41.46 (+1.05) & 39.49 \\
    \bottomrule
  \end{tabular}
  \caption{\label{tab:nb_yn_ab}
    Compared against the no-intervention baseline, Q4 system redistribution improves performance by a much larger margin in yes/no prompts as compared to A/B prompts. This is despite both involving binary choice, and especially visible in the paired accuracy metric.
  }
\end{table*}

% Q4 system redistribution improves performance by a much larger margin in yes/no prompts as compared to A/B prompts, despite both involving binary choice. This is especially visible in the paired accuracy metric.

\noindent\textbf{\textit{Whoops!}}.~Our second set of prompts is based on the Whoops!~benchmark \citep{Bitton-Guetta_2023_ICCV} and designed to simultaneously elicit both yes/no and open-ended responses.

The Whoops!~benchmark comprises AI-generated images that violate common sense. Accordingly, one of the benchmark's chief purposes is to test whether LLMs are able to detect that. To evaluate to what extent Q4 system redistribution exerts the same effect on yes/no responses on the one hand and autoregressively generated ones on the other, for each image in this benchmark we elicited: i) a yes/ no response (`\textit{Does this image depict a normal scene?}') and ii) an open-ended explanation for the response, mirroring the methodology used by \citet{parcalabescu2025do} (see Section \ref{sec:conclusion}).

This reveals cases where Q4 system redistribution changed the response from `yes' to `no' but not the semantics of the explanation. An example is shown in Table \ref{tab:whoops}, involving an image of strawberries on a pizza. In this case, although Q4 system redistribution changed the response from `yes' to ‘no’, the explanation given for the response in both cases described the image using terms such as 'not typical' and 'unconventional'.

\begin{table*}[t!]
  \centering
  \small
  \renewcommand{\arraystretch}{1.8}
  \begin{tabular}{m{0.12\textwidth} m{0.22\textwidth} m{0.30\textwidth} m{0.30\textwidth}}
    \hline
    \textbf{Prompt} & \textbf{Image} & \textbf{Q4 system redistr (prop)} & \textbf{No-intervention baseline} \\
    \hline
    Does this image show a normal scene? Explain.
    &
    \centering
    \par\vspace{5pt}
    \includegraphics[width=0.9\linewidth,keepaspectratio]{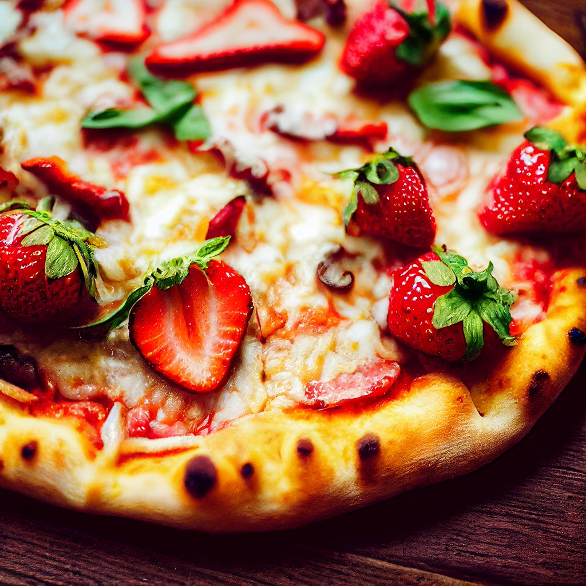}
    \par\vspace{2pt}
    &
    \textcolor{red}{\textbf{No}}, this image does not show a normal scene. The image shows a pizza with a unique and \textcolor{blue}{\textit{unusual}} topping: a large, fresh, and freshly-picked strawberry. This is \textcolor{blue}{\textit{not a typical topping}} for a pizza, as it is more commonly topped with various types of cheese, meats, and other typical toppings. The image is a close-up of the pizza, which is placed on a table, and the focus is on the unique and \textcolor{blue}{\textit{unusual}} topping.
    &
    \textcolor{red}{\textbf{Yes}}, this image shows a normal scene. It features a pizza with strawberries as toppings, placed on a wooden table. The pizza is cut into slices, and the strawberries are scattered on top of the cheese. This combination of ingredients is \textcolor{blue}{\textit{not typical}} for a traditional pizza, but it is an example of a creative and unique pizza topping. The image captures a moment where someone has prepared and served this \textcolor{blue}{\textit{unconventional}} pizza, making it a part of a regular meal or gathering.
    \\
    \hline
  \end{tabular}
  \caption{\label{tab:whoops}
    Example prompt from the Whoops! benchmark illustrating how Q4 system redistribution causes LLaVA-1.5 7B to respond `no' instead of `yes' to the same prompt, while the explanation for the response remains semantically identical under both settings.
  }
\end{table*}

This buttresses our argument that removing redundant late-layer system attention could potentially serve as a targeted means of mitigating the yes-bias.

\subsection{Results for other LLaVA architectures}
\label{app:other}
To examine how system-mediated attention imbalances are affected by differences in training regime and model scale, we perform interventions on two architectures closely related to LLaVA-1.5 7B:

\begin{itemize}
    \item \textbf{LLaVA-NeXT-(Vicuna-)7B}. Results are presented in Tables \ref{tab:prop_llavanext7b} to \ref{tab:pairwise_llavanext7b_2}. All benchmarks used for the main experiments on LLaVA-1.5 7B are included, but a subset of BEAF containing approx.~50\% of the original number of samples (14,024) needed to be used instead of the full benchmark due to memory issues. Follow-up experiments show our baseline architecture LLaVA-1.5 7B to exhibit the analogous shifts in performance with and without intervention on this subset.
    \item \textbf{LLaVA-1.5 13B}. Results are contained in Tables \ref{tab:prop_llava1513b} and \ref{tab:pairwise_llava1513b}. BEAF is not included because of memory issues analogous to those for LLaVA-NeXT.
\end{itemize}

\begin{table*}[t]
  \centering
  \small
  \renewcommand{\arraystretch}{1.8}
  \begin{tabular}{m{0.14\textwidth} m{0.12\textwidth} m{0.20\textwidth} m{0.20\textwidth} m{0.20\textwidth}}
    \hline
    \textbf{Benchmark} & \textbf{Category} & \textbf{Image} & \textbf{Positive prompt} & \textbf{Negative prompt} \\
    \hline
    \multirow{2}{=}{HallusionBench}
    &
    Chart
    &
    \centering
    \par\vspace{5pt}
    \includegraphics[width=0.9\linewidth,keepaspectratio]{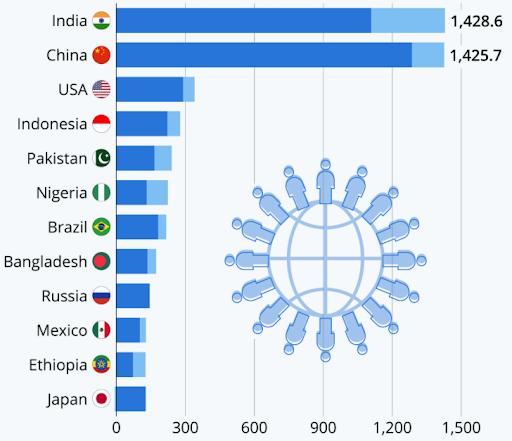}
    \par\vspace{2pt}
    &
    According to the chart, does India have the second largest population in the world?
    &
    According to the chart, does UK have the largest population in the world, followed by China and USA?
    \\
    &
    Chart
    &
    \centering
    \par\vspace{5pt}
    \includegraphics[width=0.9\linewidth,keepaspectratio]{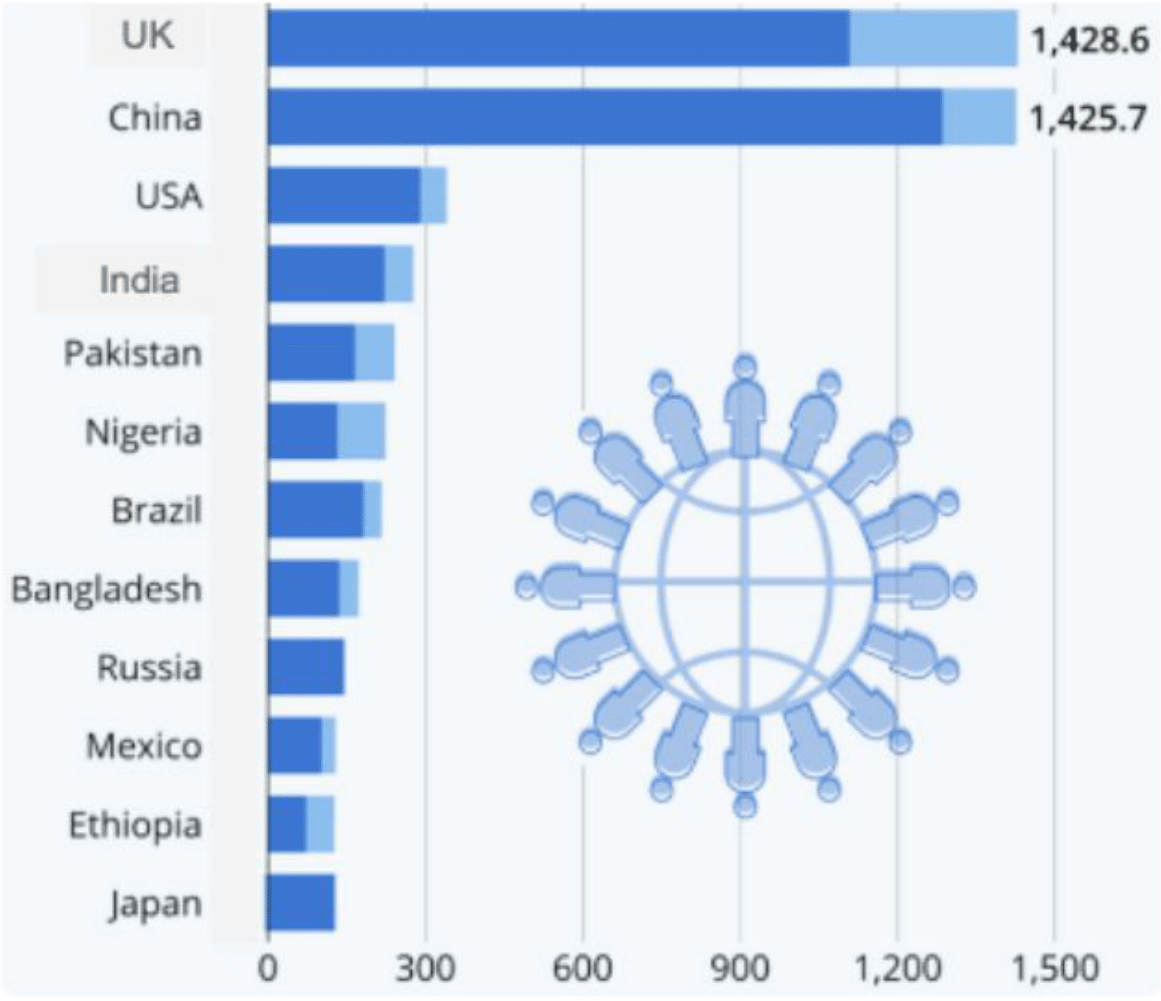}
    \par\vspace{2pt}
    &
    According to the chart, does UK have the largest population in the world, followed by China and USA?
    &
    According to the chart, does India have the second largest population in the world?
    \\
    \hline
    \multirow{2}{=}{SugarCrepe}
    &
    Add attribute
    &
    \centering
    \par\vspace{10pt}
    \includegraphics[width=0.9\linewidth,keepaspectratio]{latex/sc_000000562121.jpg}
    \par\vspace{2pt}
    &
    Two zebras are battling each other on hind legs.
    &
    Two striped-and-spotted zebras are battling each other on hind legs.
    \\
    &
    Add object
    &
    \centering
    \par\vspace{5pt}
    \includegraphics[width=0.9\linewidth,keepaspectratio]{latex/sc_000000029596.jpg}
    \par\vspace{2pt}
    &
    A nicely decorated living room and dining area.
    &
    A nicely decorated living room with a bookshelf and dining area.
    \\
    \hline
  \end{tabular}
  \caption{\label{tab:paired_prompt_examples}
    Examples from HallusionBench and SugarCrepe requiring fine-grained multimodal discrimination. Here, the correct answer hinges on minimal differences in user prompts or the image input.
  }
\end{table*}

These two architectures share LLaVA-1.5 7B's Vicuna decoder but each differ from it in one of two aspects --- LLaVA-NeXT(-Vicuna)-7B with substantially improved fine-tuning, and LLaVA-1.5 13B, which differs in scale. This reveals system-mediated attention imbalances to behave consistently across training regimes but vary more with scale.

LLaVA-NeXT 7B's improvements include more effective curation of datasets and also higher input image resolution to enhance its multimodal question-answering and reasoning capabilities \citep{liu2024llavanext}. Despite these, Tables \ref{tab:prop_llavanext7b}, \ref{tab:pairwise_llavanext7b_1} and \ref{tab:pairwise_llavanext7b_2} show that redistributing Q4 system attention both proportionally and in a pairwise fashion yields qualitatively similar suppression of yes-bias across most benchmarks. These findings indicate on the whole that late-layer system-mediated attention imbalances are not meaningfully reduced by improved image processing or training alone. Subtle differences from LLaVA-1.5 7B, most notably in system-to-text and system-to-image transfers being better matched in their performance, suggest that these improvements partly address image attention deficits, but without resolving the underlying redundant system attention.

Scale appears to have a larger impact on attention imbalances. Despite lacking LLaVA-NeXT's enhancements, LLaVA-1.5 13B achieves stronger baseline performance in most benchmarks, surpassing even the best interventions on both 7B models (Table \ref{tab:prop_llava1513b}). In this regime, Q4 system redistribution no longer produces consistent gains and can in many cases worsen the yes-bias. Nonetheless, targeted system-to-image and system-to-text redistributions still yield approx.~9-20\% improvements in simple accuracy in HallusionBench and SugarCrepe respectively (Table \ref{tab:pairwise_llava1513b}). This indicates that system-mediated attention imbalances remain potentially useful levers against hallucination at larger scales, but assume different characteristics which warrant further investigation. 

\begin{table*}
\centering
\footnotesize
\renewcommand{\arraystretch}{1.15}
\begin{tabular}{
p{0.25\textwidth}
p{0.12\textwidth}
p{0.15\textwidth}
p{0.15\textwidth}
}
\toprule
\textbf{Benchmark (\% prompts with ground-truth yes)} & \textbf{Metric} & \textbf{Q4 system redistr (prop)} & \textbf{No-intervention baseline} \\
\midrule

\multirow{3}{=}{BEAF (sample) \\ (33.71\% yes)}
& \makecell[l]{Simple acc$^{\Uparrow}$}
& \makecell[l]{\textbf{90.67} (+0.82)}
& 89.93 \\

& \makecell[l]{Paired acc$^{\Uparrow}$}
& \makecell[l]{\textbf{87.72} (+1.07)}
& 86.79 \\

& \makecell[l]{Yes-rate$^{\Delta\downarrow}$}
& \makecell[l]{31.50 (–6.55)}
& \makecell[l]{\textbf{35.34} (+4.83)} \\

\midrule
\multirow{3}{=}{HallusionBench \\ (42.17\% yes)}
& \makecell[l]{Simple acc$^{\Uparrow}$}
& \makecell[l]{\textbf{52.89} (+11.04)}
& 47.63 \\

& \makecell[l]{Paired acc$^{\Uparrow}$}
& \makecell[l]{\textbf{16.74} (+35.85)}
& 12.33 \\

& \makecell[l]{Yes-rate$^{\Delta\downarrow}$}
& \makecell[l]{\textbf{61.51} (+45.87)}
& \makecell[l]{87.17 (+106.71)} \\

\midrule
\multirow{3}{=}{MME \\ (50.00\% yes)}
& \makecell[l]{Simple acc$^{\Uparrow}$}
& \makecell[l]{40.61 (–46.36)}
& \textbf{75.70} \\

& \makecell[l]{Paired acc$^{\Uparrow}$}
& \makecell[l]{0.77 (–95.00)}
& \textbf{15.38} \\

& \makecell[l]{Yes-rate$^{\Delta\downarrow}$}
& \makecell[l]{20.56 (–58.89)}
& \makecell[l]{\textbf{56.44} (+12.89)} \\

\midrule
\multirow{3}{=}{NaturalBench \\ (50.00\% yes)}
& \makecell[l]{Simple acc$^{\Uparrow}$}
& \makecell[l]{\textbf{67.81} (+4.05)}
& 65.17 \\

& \makecell[l]{Paired acc$^{\Uparrow}$}
& \makecell[l]{\textbf{37.86} (+19.22)}
& 31.76 \\

& \makecell[l]{Yes-rate$^{\Delta\downarrow}$}
& \makecell[l]{\textbf{65.74} (+31.48)}
& \makecell[l]{77.17 (+54.34)} \\

\midrule
\multirow{3}{=}{SugarCrepe \\ (50.00\% yes)}
& \makecell[l]{Simple acc$^{\Uparrow}$}
& \makecell[l]{\textbf{64.56} (+4.35)}
& 61.87 \\

& \makecell[l]{Paired acc$^{\Uparrow}$}
& \makecell[l]{\textbf{29.23} (+22.58)}
& 23.85 \\

& \makecell[l]{Yes-rate$^{\Delta\downarrow}$}
& \makecell[l]{\textbf{83.24} (+66.48)}
& \makecell[l]{86.81 (+73.63)} \\

\midrule
\multirow{3}{=}{Winoground \\ (50.00\% yes)}
& \makecell[l]{Simple acc$^{\Uparrow}$}
& \makecell[l]{\textbf{58.81} (+6.21)}
& 55.38 \\

& \makecell[l]{Paired acc$^{\Uparrow}$}
& \makecell[l]{\textbf{21.50} (+70.30)}
& 12.63 \\

& \makecell[l]{Yes-rate$^{\Delta\downarrow}$}
& \makecell[l]{\textbf{76.81} (+53.63)}
& \makecell[l]{91.25 (+82.50)} \\

\bottomrule
\end{tabular}
\caption{\label{tab:prop_llavanext7b} Changes in simple accuracy, paired accuracy and yes-rate across six paired-prompt benchmarks for LLaVA-NeXT-Vicuna-7B, with and without proportional Q4 system redistribution. Like for LLaVA-1.5 7B, the intervention yields gains in all benchmarks but MME. Best results for each metric are in bold. For simple and paired accuracy, percentage differences relative to the no-intervention baseline are shown in brackets. The differences for yes-rate are with reference to the actual proportion of prompts with ground-truth `yes', shown bracketed in the Benchmark column --- the best yes-rate is the one that deviates least from that ground-truth value.}
\end{table*}

\begin{table*}[h]
\centering
\footnotesize
\begin{tabular}{l|ccc|ccc|ccc|ccc}
\toprule
\multirow{3}{*}{\textbf{Source}} & \multicolumn{12}{c}{\textbf{Recipient}} \\
\cline{2-13}
& \multicolumn{3}{c}{\textbf{HallusionBench}} 
& \multicolumn{3}{c}{\textbf{NaturalBench (yes/no)}} 
& \multicolumn{3}{c}{\textbf{SugarCrepe}} 
& \multicolumn{3}{c}{\textbf{Winoground}} \\
\cline{2-13}
& \textit{System} & \textit{Image} & \textit{Text}
& \textit{System} & \textit{Image} & \textit{Text}
& \textit{System} & \textit{Image} & \textit{Text}
& \textit{System} & \textit{Image} & \textit{Text} \\
\midrule

\textit{System}
& \cellcolor{lightgray}47.63 & \underline{58.15} & \textbf{58.36}
& \cellcolor{lightgray}65.17 & \textbf{69.93} & 
\underline{69.84}
& \cellcolor{lightgray}61.87 & \underline{67.58} & \textbf{67.69}
& \cellcolor{lightgray}55.36 & \textbf{61.12} & \underline{61.00} \\[1ex]

\textit{Image}
& 47.53 & \cellcolor{lightgray}47.63 & 47.42
& 65.17 & \cellcolor{lightgray}65.17 & 65.21
& 61.98 & \cellcolor{lightgray}61.87 & 61.87
& 55.19 & \cellcolor{lightgray}55.36 & 55.31 \\[1ex]

\textit{Text}
& 50.37 & 50.58 & \cellcolor{lightgray}47.63
& 67.78 & 66.67 & \cellcolor{lightgray}65.17
& 64.23 & 64.29 & \cellcolor{lightgray}61.87
& 57.88 & 57.50 & \cellcolor{lightgray}55.36 \\[1ex]

\bottomrule
\end{tabular}
\caption{Changes in simple accuracy in HallusionBench, NaturalBench (yes/no), SugarCrepe, and Winoground under pairwise attention-weight transfers between all combinations of source and recipient modalities in Q4 (layers 25-32) of LLaVA-NeXT-Vicuna 7B. Although transfers with system as the source modality consistently yield the biggest improvements in performance, system-to-text and system-to-image transfers perform more similarly to each other than in LLaVA-1.5 7B.}
\label{tab:pairwise_llavanext7b_1}
\end{table*}

\begin{table*}[h]
\centering
\footnotesize
\begin{tabular}{l|ccc|ccc}
\toprule
\multirow{3}{*}{\textbf{Source}} & \multicolumn{6}{c}{\textbf{Recipient}} \\
\cline{2-7}
& \multicolumn{3}{c}{\textbf{BEAF}} 
& \multicolumn{3}{c}{\textbf{MME}} \\
\cline{2-7}
& \textit{System} & \textit{Image} & \textit{Text}
& \textit{System} & \textit{Image} & \textit{Text} \\
\midrule

\textit{System}
& \cellcolor{lightgray}89.93 & 89.28 & 89.33
& \cellcolor{lightgray}75.70 & 61.58 & 64.20 \\[1ex]

\textit{Image}
& 89.92 & \cellcolor{lightgray}89.93 & 89.97
& 75.74 & \cellcolor{lightgray}75.70 & \underline{75.82} \\[1ex]

\textit{Text}
& \underline{90.68} & \textbf{90.84} & \cellcolor{lightgray}89.93
& \textbf{76.16} & 75.06 & \cellcolor{lightgray}75.70 \\[1ex]

\bottomrule
\end{tabular}
\caption{Changes in simple accuracy in BEAF and MME under pairwise attention-weight transfers between all combinations of source and recipient modalities in Q4 (layers 25-32) of LLaVA-NeXT-Vicuna 7B. Transfers with system as the source modality consistently underperform the no-intervention baseline (shaded).}
\label{tab:pairwise_llavanext7b_2}
\end{table*}

\begin{table*}
\centering
\footnotesize
\renewcommand{\arraystretch}{1.15}
\begin{tabular}{
p{0.25\textwidth}
p{0.12\textwidth}
p{0.15\textwidth}
p{0.15\textwidth}
}
\toprule
\textbf{Benchmark (\% prompts with ground-truth yes)} &
\textbf{Metric} &
\textbf{Q4 system redistr (prop)} &
\textbf{No-intervention baseline} \\
\midrule

\multirow{3}{=}{HallusionBench \\ (42.17\% yes)}
& \makecell[l]{Simple acc$^{\Uparrow}$}
& \makecell[l]{\textbf{51.31} (+0.83)}
& 50.89 \\

& \makecell[l]{Paired acc$^{\Uparrow}$}
& \makecell[l]{15.58 (–1.47)}
& \textbf{15.81} \\

& \makecell[l]{Yes-rate$^{\Delta\downarrow}$}
& \makecell[l]{\textbf{65.62} (+55.60)}
& \makecell[l]{67.93 (+61.08)} \\

\midrule

\multirow{3}{=}{MME \\ (50.00\% yes)}
& \makecell[l]{Simple acc$^{\Uparrow}$}
& \makecell[l]{79.32 (–0.05)}
& \textbf{79.36} \\

& \makecell[l]{Paired acc$^{\Uparrow}$}
& \makecell[l]{27.69 (–14.29)}
& \textbf{32.31} \\

& \makecell[l]{Yes-rate$^{\Delta\downarrow}$}
& \makecell[l]{54.93 (+9.86)}
& \makecell[l]{\textbf{47.89} (–4.21)} \\

\midrule

\multirow{3}{=}{NaturalBench \\ (50.00\% yes)}
& \makecell[l]{Simple acc$^{\Uparrow}$}
& \makecell[l]{68.36 (–1.59)}
& \textbf{69.46} \\

& \makecell[l]{Paired acc$^{\Uparrow}$}
& \makecell[l]{39.05 (–4.57)}
& \textbf{40.92} \\

& \makecell[l]{Yes-rate$^{\Delta\downarrow}$}
& \makecell[l]{65.83 (+31.66)}
& \makecell[l]{\textbf{61.25} (+22.50)} \\

\midrule

\multirow{3}{=}{SugarCrepe \\ (50.00\% yes)}
& \makecell[l]{Simple acc$^{\Uparrow}$}
& \makecell[l]{\textbf{68.96} (+0.88)}
& 68.35 \\

& \makecell[l]{Paired acc$^{\Uparrow}$}
& \makecell[l]{\textbf{38.46} (+3.24)}
& 37.25 \\

& \makecell[l]{Yes-rate$^{\Delta\downarrow}$}
& \makecell[l]{\textbf{76.54} (+53.08)}
& \makecell[l]{77.36 (+54.73)} \\

\midrule

\multirow{3}{=}{Winoground \\ (50.00\% yes)}
& \makecell[l]{Simple acc$^{\Uparrow}$}
& \makecell[l]{58.56 (–2.19)}
& \textbf{59.88} \\

& \makecell[l]{Paired acc$^{\Uparrow}$}
& \makecell[l]{21.50 (–12.69)}
& \textbf{24.63} \\

& \makecell[l]{Yes-rate$^{\Delta\downarrow}$}
& \makecell[l]{79.31 (+58.63)}
& \makecell[l]{\textbf{74.38} (+48.75)} \\

\bottomrule
\end{tabular}
\caption{\label{tab:prop_llava1513b} Changes in simple accuracy, paired accuracy and yes-rate across five paired-prompt benchmarks for LLaVA-1.5 13B, with and without proportional Q4 system redistribution. Compared to LLaVA-1.5 7B, the intervention produces very limited gains and increases the model's yes-rate for three benchmarks (MME, NaturalBench, Winoground). Best results for each metric are in bold. For simple and paired accuracy, percentage differences relative to the no-intervention baseline are shown in brackets. The differences for yes-rate are with reference to the actual proportion of prompts with ground-truth `yes', shown bracketed in the Benchmark column --- the best yes-rate is the one that deviates least from that ground-truth value.}
\end{table*}

\begin{table*}[h]
\centering
\footnotesize
\begin{tabular}{l|ccc|ccc}
\toprule
\multirow{3}{*}{\textbf{Source}} & \multicolumn{6}{c}{\textbf{Recipient}} \\
\cline{2-7}
& \multicolumn{3}{c}{\textbf{HallusionBench}} 
& \multicolumn{3}{c}{\textbf{SugarCrepe}} \\
\cline{2-7}
& \textit{System} & \textit{Image} & \textit{Text}
& \textit{System} & \textit{Image} & \textit{Text} \\
\midrule

\textit{System}
& \cellcolor{lightgray}50.89 & \textbf{61.30} & \underline{60.99}
& \cellcolor{lightgray}68.35 & \underline{74.40} & \textbf{74.62} \\[1ex]

\textit{Image}
& 51.63 & \cellcolor{lightgray}50.89 & 51.21
& 68.46 & \cellcolor{lightgray}68.35 & 68.35 \\[1ex]

\textit{Text}
& 49.32 & 50.68 & \cellcolor{lightgray}50.89
& 68.35 & 68.79 & \cellcolor{lightgray}68.35 \\[1ex]

\bottomrule
\end{tabular}
\caption{Changes in simple accuracy in HallusionBench and SugarCrepe under pairwise attention-weight transfers between all combinations of source and recipient modalities in Q4 (layers 25-32) of LLaVA-1.5 13B. Transfers with system as the source modality consistently underperform the no-intervention baseline (shaded).}
\label{tab:pairwise_llava1513b}
\end{table*}

\subsection{Results for Qwen2-VL-7B-Instruct}
\label{app:qwen2}
For the purposes of cross-architectural comparison, the Qwen family of models is especially interesting because \citet{luo2026from} note Qwen2.5-VL-Instruct to not be very receptive to training-free interventions of the kind used in this study. We speculate that that may be due to, among various other factors, Qwen's joint training of the image encoder and text decoder leading to more efficient allocation of attention weights in the text decoder as compared to LLaVA-family architectures without joint training \citep{team2024qwen2}.

Indeed, as shown in Table \ref{tab:qwen2}, Q4 system redistribution only benefits performance negligibly in Qwen2-VL-7B-Instruct, never by more than 0.5\%. Only simple accuracy is reported.

\begin{table*}[h]
\centering
\footnotesize
\begin{tabular}{lccc}
\toprule
\textbf{Benchmark} & \textbf{Q4 system redistr (prop)} & \textbf{Q4 text-to-image redistr} & \textbf{No-intervention baseline} \\
\midrule
BEAF & \textbf{89.16} (+0.01) & 86.08 (-3.44) & 89.15 \\
HallusionBench & \textbf{69.36} (+0.37) & 68.59 (-0.74) & 69.10 \\
MME            & 86.41 (-0.17) & 86.22 (-0.39) & \textbf{86.56} \\
NaturalBench   & \textbf{74.90} (+0.20) & 74.67 (-0.11) & 74.75 \\
SugarCrepe     & \textbf{77.31} (+0.29) & 72.69 (-5.71) & 77.09 \\
Winoground     & \textbf{75.00} (0.00) & 73.83 (-1.56) & \textbf{75.00} \\
\bottomrule
\end{tabular}
\caption{Simple accuracy of Qwen2-VL-7B-Instruct in six benchmarks under two different intervention settings, alongside the no-intervention baseline. Interventions here impact much less considerably on performance as compared to the LLaVA-family models. Nevertheless, in accordance with this study's system-mediated hypothesis, proportionally redistributing Q4 system attention (Q4 system redistr (prop)) is consistently less harmful than Q4 text-to-image transfer, demonstrating the former's superiority as an empirical account. This is despite the performance differences arising under intervention being much smaller than in the LLaVA-family models shown previously. Best results per benchmark bolded.}
\label{tab:qwen2}
\end{table*}

Nonetheless, as the same table shows, the results broadly agree with our findings for LLaVA-1.5 7B, providing this study's core arguments with compelling support across model families:
\begin{enumerate}
    \item Proportional Q4 system redistribution, an operationalisation of the system-mediated hypothesis, is consistently more beneficial than Q4 text-to-image redistribution, which represents the image-centric hypothesis (see Sections \ref{sec:intro} and \ref{sec:pairwise})
    \item Q4 system redistribution brings about a dip in performance for MME but not for the other datasets
\end{enumerate}

The first point crucially demonstrates that our system-mediated account is superior to the prevailing image-centric alternative for both the LLaVA-family models tested and Qwen2-VL-7B-Instruct.

The second point lends credence to our hypothesis in Section \ref{sec:error_analysis} that the presence of late-layer system attention encourages the use of coarse-grained aggregated input representations, benefiting tasks such as MME. Performance in this benchmark is therefore harmed when this attention is redistributed, albeit by a much smaller margin where Qwen2-VL-Instruct is concerned, potentially for the reasons mentioned at the start of this appendix.

\end{document}